\documentclass{article} %
\usepackage{iclr2024_conference,times}

\usepackage{amsmath,amsfonts,bm}

\def\eqref#1{equation~\ref{#1}}

\def\1{\bm{1}}

\def\mT{{\bm{T}}}

\DeclareMathAlphabet{\mathsfit}{\encodingdefault}{\sfdefault}{m}{sl}
\SetMathAlphabet{\mathsfit}{bold}{\encodingdefault}{\sfdefault}{bx}{n}

\usepackage{url}
\usepackage{ftnxtra}
\usepackage{lineno}
\usepackage{amsmath}
\usepackage{amssymb}
\usepackage{booktabs}
\usepackage{graphicx}
\usepackage{multirow}
\usepackage{cases}
\usepackage{array}
\usepackage{threeparttable}
\usepackage{comment}
\usepackage{amsmath,amssymb}
\usepackage{xspace}
\usepackage{makecell}
\usepackage{colortbl}
\usepackage{mathrsfs}
\usepackage{colortbl}
\usepackage{rotating}
\usepackage{siunitx}
\usepackage{pifont}
\usepackage{bm}
\usepackage{xcolor}
\usepackage[export]{adjustbox}
\usepackage{algpseudocode}
\usepackage{algorithm}
\usepackage{enumitem}
\usepackage[misc]{ifsym}
\usepackage[bookmarks=false]{hyperref}

\usepackage{wrapfig,lipsum,booktabs}

\title{DV-3DLane: End-to-end Multi-modal 3D Lane Detection with Dual-view Representation}

\author{Yueru Luo\textsuperscript{\,\rm 1,2}, 
      \, Shuguang Cui\textsuperscript{\,\rm 2,1},
      \, Zhen Li\textsuperscript{\,\rm 2,1, \Letter} \\
\textsuperscript{\rm 1} FNii, CUHK-Shenzhen \quad
\textsuperscript{\rm 2} School of Science and Engineering, CUHK-Shenzhen \quad \\
\texttt{\{222010057@link.,shuguangcui@,lizhen@\}cuhk.edu.cn}
}

\newcommand{\near}{\textit{near\xspace}}
\newcommand{\far}{\textit{far\xspace}}

\newcommand{\uv}{PV\xspace}
\newcommand{\bev}{BEV\xspace}
\newcommand{\uvfull}{perspective-view\xspace}

\definecolor{impblue}{rgb}{0.0, 0.5, 0.8}
\definecolor{mygray}{gray}{0.9}
\definecolor{Green}{rgb}{0.26, 0.62, 0.305}

\definecolor{brw}{rgb}{0.72941176, 0.43921569, 0.30980392}
\definecolor{noise}{rgb}{0.51764706, 0.36862745, 0.76078431}
\definecolor{gt}{rgb}{0.8,0.15,0.15}
\definecolor{prev_sota}{rgb}{0.91,0.69,0.04}

\newcommand{\wrt}{\textit{w}.\textit{r}.\textit{t}. }

\usepackage[capitalize]{cleveref}
\crefname{section}{Sec.}{Secs.}
\Crefname{section}{Section}{Sections}
\Crefname{table}{Table}{Tables}
\crefname{table}{Tab.}{Tabs.}

\makeatletter
\DeclareRobustCommand\onedot{\futurelet\@let@token\@onedot}
\def\@onedot{\ifx\@let@token.\else.\null\fi\xspace}
\def\eg{\emph{e.g}\onedot} 
\def\ie{\emph{i.e}\onedot}

\makeatother

\newcommand{\modelname}[0]{DV-3DLane\xspace}

\iclrfinalcopy %
\begin{document}

\maketitle

\let\thefootnote\relax\footnotetext{$^\textrm{\Letter}$ Corresponding author.}

\begin{abstract}
Accurate 3D lane estimation is crucial for ensuring safety in autonomous driving. 
However, prevailing monocular techniques suffer from depth loss and lighting variations, hampering accurate 3D lane detection. 
In contrast, LiDAR points offer geometric cues and enable precise localization. 
In this paper, we present~\modelname, a novel end-to-end {\textbf{D}ual}-{\textbf{V}iew} multi-modal \textbf{3D} \textbf{Lane} detection framework that synergizes the strengths of both images and LiDAR points. 
We propose to learn multi-modal features in dual-view spaces, \ie, \textit{perspective view} (\uv) and \textit{bird's-eye-view} (\bev), effectively leveraging the modal-specific information.
To achieve this, we introduce three designs:
\textbf{1)} A bidirectional feature fusion strategy that integrates multi-modal features into each view space, exploiting their unique strengths.
\textbf{2)} A unified query generation approach that leverages lane-aware knowledge from both \uv and \bev spaces to generate queries. 
\textbf{3)} A 3D dual-view deformable attention mechanism, which aggregates discriminative features from both \uv and \bev spaces into queries for accurate 3D lane detection.  
Extensive experiments on the public benchmark, OpenLane, demonstrate the efficacy and efficiency of~\modelname. It achieves state-of-the-art performance, with a remarkable \textbf{11.2} gain in F1 score and a substantial \textbf{53.5\%} reduction in errors. 
The code is available at \url{https://github.com/JMoonr/dv-3dlane}.
\end{abstract}

\section{Introduction}

\begin{wrapfigure}{r}{0.5\linewidth}
    \centering
    \vspace{-13mm}
        \includegraphics[width=1\linewidth]{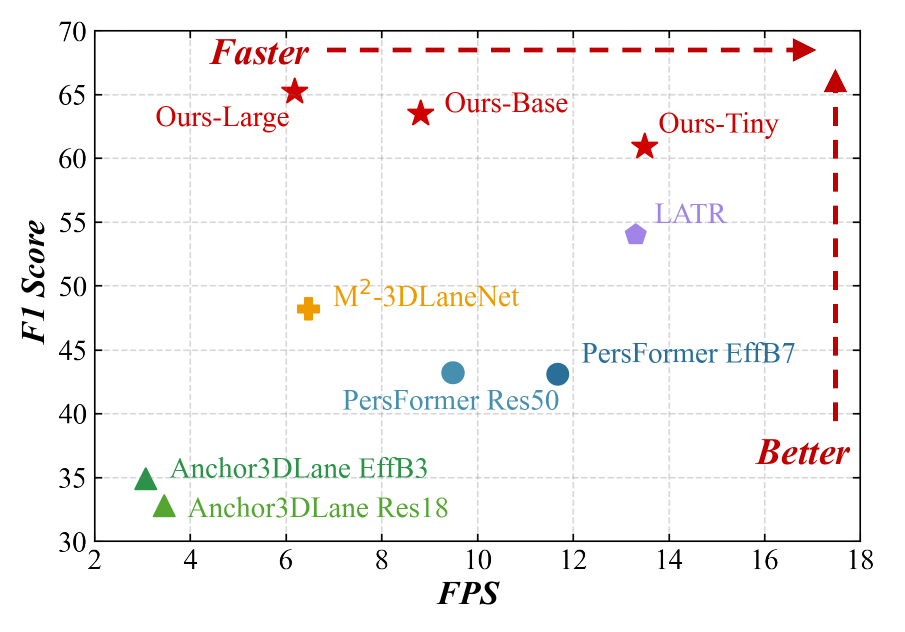}
        \vspace{-8mm}
        \caption{\textbf{FPS vs.\ F1 score.} All models are tested on a single V100 GPU, and F1-score is evaluated with a harsh distance threshold of \textbf{0.5m} on the OpenLane-1K dataset. Our model sets a new state-of-the-art, and our tiny version surpasses all previous methods
        with the fastest FPS. More details can be found in~\Cref{tab:main_results} and the Appendix.}
        \label{fig:fps}
        \vspace{-4mm}
\end{wrapfigure}

Autonomous driving (AD) technology in recent years has made remarkable strides, bringing us closer to the realization of fully self-driving vehicles. Within this field, one of the key challenges is the accurate detection of 3D lanes, a critical component for ensuring safe and reliable navigation. 3D lane detection entails identifying the 3D positions of lane boundaries in the environment, providing essential data for tasks like path planning and vehicle control.

3D lane detection is proposed to mitigate the limitations posed by the absence of depth information in 2D prediction.
Currently, the majority of 3D lane detection methods rely on vision-centric approaches, \ie, monocular solutions, where some designs are naturally borrowed and benefit from advances in 2D lane methods. Taking the \uvfull (\uv) image as input, these monocular methods mainly utilize the inverse perspective mapping (IPM)~\cite{mallot1991inverse} technique to warp the \uv features into \bev.
However, there are misalignment issues in the IPM-based methods when encountering non-flat roads, due to the rigid flat assumption of IPM~\cite{nedevschi20043d, yan2022once}.
While some recent efforts have been made to address this issue and have shown promising results by directly predicting 3D lanes in \uv~\cite{bai2022curveformer, huang2023anchor3dlane, luo2023latr}, these monocular 3D approaches, as vision-centric solutions, inevitably get stuck in capturing the complexity of real-world driving scenarios, when encountering adverse weather and lighting conditions.
In contrast, as an active sensor, LiDAR excels in spatial localization and 3D structure perception, complementing the capabilities of passive sensor cameras, and it gets more widely used thanks to hardware advancements.
A bunch of recent works in 3D object detection have demonstrated the power of LiDARs~\cite{zhou2018voxelnet, lang2019pointpillars, yin2021center} and multiple modalities~\cite{liang2019multi, wang2021pointaugmenting,yang2022deepinteraction, li2022deepfusion, chen2023futr3d} in autonomous driving scenarios. Whereas, \textit{fewer endeavors}~\cite{bai2018deep,luo2022m} \textit{have been made to exploit multi-modal strength for 3D lane detection}. 
Albeit using extra LiDAR data, M$^2$-3DLane~\cite{luo2022m} failed to make full use of features in image space which is crucial to 3D lane performance. Besides,  M$^2$-3DLane employs a naive fusion to aggregate multi-modal features, resulting in inferior performance to the camera-only methods(\eg,~\cite{luo2023latr}).

Given the rich semantics inherent in images and the accurate positional information afforded by the BEV representation~\cite{philion2020lift, li2022bevformer}, we strive to exploit the multi-modal features to enhance the performance of 3D lane detection.
Existing methods tend to fuse two modalities into a \textit{single} space~\cite{liang2022bevfusion, liu2023bevfusion}, \eg, \bev, for feature extraction and subsequent prediction.
However,
this approach constrains the model's capacity to harness modality-specific features.
We contend that features represented in \textit{both \uv space and \bev space bear significance}, facilitating improved representation learning. 
Motivated by the above observation, we introduce \textbf{\modelname}, a novel end-to-end multi-modal 3D lane detection framework.

To maintain a dual-view space representation, we adopt a symmetric backbone consisting of a \uv branch and a \bev branch to extract features in \uv and \bev spaces, respectively.
To leverage the merits of both images and points for comprehensive feature learning in each view, we design a \textit{bidirectional} feature fusion (BFF) strategy.
Subsequently, to effectively facilitate query-based detection using the retained dual-view features, we devise a \textit{unified} query generator (UQG).
This generator initially produces two sets of lane-aware queries: one from the \uv space and the other from the \bev space. 
{These two query sets are compelled to capture lane knowledge regarding semantics and spatiality, guided by auxiliary 2D segmentation supervision.} 
Further, these two sets are then combined into a unified set that serves the decoder.
To achieve the unification of dual-view queries, we propose a \textit{lane-centric} clustering technique.
Besides, we employ a Transformer decoder to aggressively integrate discriminative features from both views into the unified queries.
For effective feature aggregation across different view spaces, we introduce a 3D dual-view deformable attention mechanism that considers the inherent properties of 3D space, resulting in deformed 3D sample points.
These 3D sample points are then projected onto the \uv and \bev planes, yielding 2D sample points in each respective view space. 
These projected 2D points are utilized for feature sampling within their respective view spaces.

In summary, our contributions are threefold :
\begin{itemize}[itemsep=1mm,leftmargin=4mm]
\vspace{-2mm}
    \item We introduce~\modelname, an end-to-end multi-modal 3D lane detection framework that harnesses the power of dual-view representation.
    \item We devise the BFF strategy to mutually fuse features across modalities, and design the UQG to merge lane-aware queries
    from dual views, yielding a unified query set. Further, a 3D dual-view deformation attention mechanism is introduced to aggregate dual-view features effectively.
    \item We conduct thorough experiments on the OpenLane benchmark to validate the effectiveness of our method. Experimental results show that~\modelname surpasses previous methods significantly, achieving an impressive {\textbf{11.2}-point} gain in F1 score and a remarkable {\textbf{{53.5\%} reduction}} in errors (on average).
Moreover, a 3D dual-view deformation attention mechanism is introduced to aggregate dual-view features effectively.
\end{itemize}

\section{Related Work}

\subsection{2D Lane Detection}
Recent works in 2D lane detection can be broadly categorized into four main approaches:
\textbf{1)} Segmentation-based methods~\cite{lee2017vpgnet, pan2017spatial, neven2018towards, hou2019learning, xu2020curvelane, zheng2021resa} devote to classifying pixels into lanes or the background, necessitating further post-processing steps (\eg, grouping and curve fitting) to produce lane instances.
\textbf{2)} Anchor-based methods, inspired by region-based object detectors such as Faster-RCNN~\cite{ren2015faster}, employ line-like anchors to localize lanes~\cite{wang2018lanenet, li2019line, tabelini2021keep}. To overcome the limitations of straight-line constraints,~\cite{jin2022eigenlanes} employ eigenlane space to produce diverse lane shape candidates.
\textbf{3)} Point-based methods~\cite{ko2021key, qu2021focus, wang2022keypoint, xu2022rclane} attempt to flexibly localize key points along each lane instance and subsequently group the points belonging to the same lane.
\textbf{4)} Parametric methods~\cite{van2019end, tabelini2021polylanenet, liu2021end, feng2022rethinking} formulate lane detection as a curve fitting problem, leveraging prior knowledge about lane shapes by representing them using various parametric forms, such as polynomials and splines.

\begin{figure*}[th]
\centering
   \includegraphics[width=1.0\linewidth]{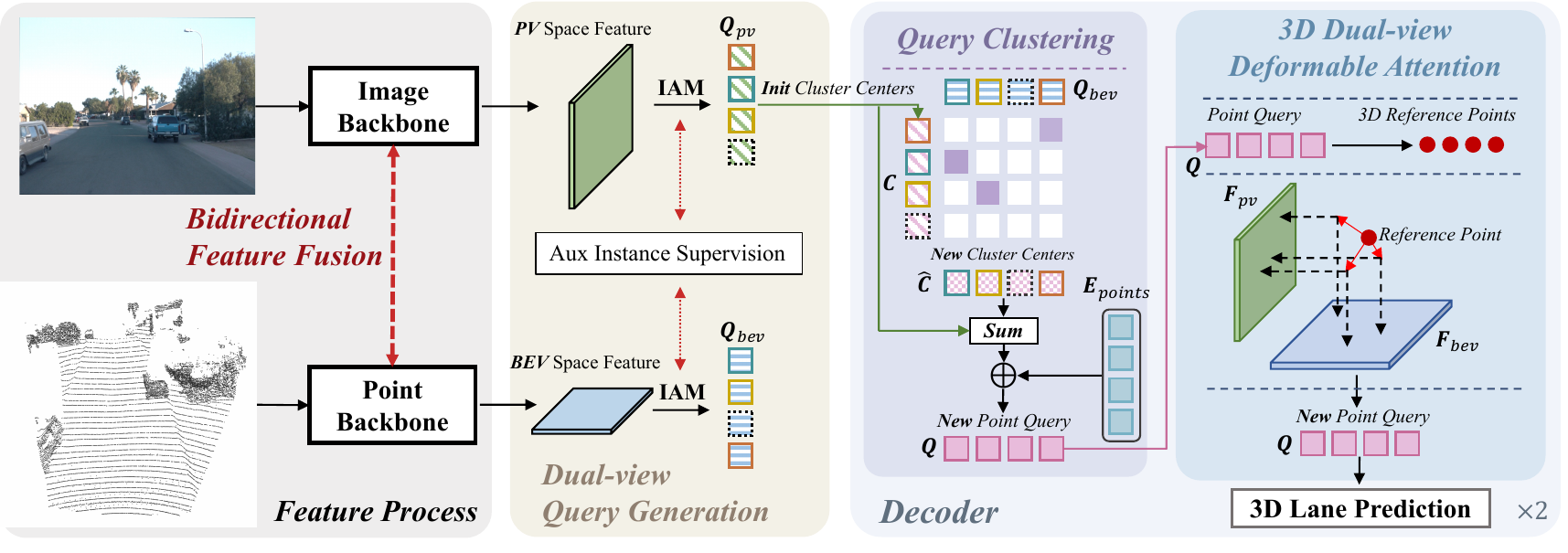} 
   \vspace{-5mm}
\caption{\textbf{Overview of~\modelname.} First, images and point clouds undergo separate processing by the image backbone and point backbone.
In the middle stage of backbones, we introduce Bidirectional Feature Fusion (BFF) to fuse multi-modal features across views. 
Subsequently, the instance activation map (IAM) is utilized to produce lane-aware queries $\mathbf{Q}_{pv}$ and $\mathbf{Q}_{bev}$.
These queries are then subjected to Dual-view Query Clustering, which aggregates dual-view query sets $\mathbf{Q}_{pv}$ and $\mathbf{Q}_{bev}$ into a unified query set $\mathbf{C}$, further augmented with learnable point embeddings $\mathbf{E}_{points}$ to form query $\mathbf{Q}$. 
Additionally, we introduce 3D Dual-view Deformable Attention to consistently aggregate point features from both view features $\mathbf{F}_{pv}$ and $\mathbf{F}_{bev}$ into $\mathbf{Q}$.
$\oplus$ denotes broadcast summation.
Notably, the $\oplus \, \mathbf{E}_{points}$ operation is performed only in the first layer, while in the following layer, $\oplus \, \mathbf{Q}$ is utilized.
Different colored boxes~\includegraphics[height=0.85em]{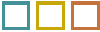} denote queries targeting different lanes; dashed boxes~\includegraphics[height=0.8em]{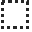} represent the background, and box texture indicates features.
}
\label{fig:fig2}
\vspace{-4mm}
\end{figure*}

\subsection{3D Lane Detection}
\label{sec:related_3d}
Existing methods center on vision-centric solutions and draw inspiration from the 2D task. Typically, monocular approaches~\cite{garnett20193d, efrat20203d, guo2020gen, chen2022persformer, wang2023bev,liu2022learning, li2022reconstruct, ai2023ws, yao2023sparse} construct surrogate representations using inverse perspective mapping (IPM), and perform predictions in this surrogate space.
Nonetheless, due to its planar assumption, IPM inherently introduces discrepancies between the perspective and the surrogate view in non-flat areas. 
To address this limitation, recent efforts have endeavored to predict 3D lanes from the perspective view~\cite{yan2022once, bai2022curveformer, huang2023anchor3dlane, luo2023latr}, or employ a depth-aware projection to enhance lane perception by incorporating LiDAR information~\cite{luo2022m}.

\subsection{Multi-modal Detection}
Despite advancements in lane detection, multi-modal methods remain relatively underexplored. 
Previous works typically utilize either \bev~\cite{bai2018deep, yin2020fusionlane, luo2022m} or \uv~\cite{zhang2021channel} as representation spaces for performing 2D lane segmentation~\cite{yin2020fusionlane, zhang2021channel} or 3D lane detection~\cite{bai2018deep, luo2022m}.
For \bev-based methods, ~\cite{bai2018deep} rasterizes LiDAR points to create a BEV image and transforms \uv images into \bev using the estimated ground height derived from the LiDAR data. 
Similarly, M$^2$-3DLane~\cite{luo2022m} utilizes the BEV space to fuse multi-modal features.
To project \uv features into \bev space, they lift compact 2D features into 3D space guided by the depth map and further employ a pillar-based method~\cite{lang2019pointpillars} to splat them into BEV. 
While these methods primarily focus on 3D tasks, ~\cite{yin2020fusionlane} leverages BEV space for fusing camera and LiDAR features, serving for 2D BEV lane segmentation.
Conversely, ~\cite{zhang2021channel} adopts \uv to fuse multi-modal features for 2D lane segmentation.
In contrast to lane detection, multi-modal methods have been extensively studied in 3D object detection, with most previous multi-modal methods attempting to fuse image features into BEV space due to its compactness and interoperability for ambient perception~\cite{ma2022vision}. 
These methods either adopt point-level fusion~\cite{sindagi2019mvx,wang2021pointaugmenting,yin2021multimodal} to paint points, instance-level fusion to project 3D proposals to image space~\cite{yoo20203d,bai2022transfusion}, or feature-level fusion to transform features from \uv space into \bev space~\cite{liu2023bevfusion,liang2022bevfusion}. 
However, few works consider both the perspective view and BEV simultaneously.

\section{Methodology}
The overall framework of our~\modelname is depicted in~\Cref{fig:fig2}.
~\Cref{sec:featfusion} describes the bidirectional feature fusion module, which merges different modalities bidirectionally and constructs multi-modal features in both \uv and \bev spaces.
In~\Cref{sec:genquery}, we present the unified query generator, which generates two lane-aware query sets from dual views and unifies them into a shared space in a lane-centric manner.
~\Cref{sec:dualattn} introduces the 3D dual-view deformable attention module, which effectively aggregates dual-view features into unified queries, serving for prediction.

\subsection{Bidirectional Feature Fusion}
\label{sec:featfusion}

\begin{figure}[ht]
\centering
\vspace{-5mm}
\begin{minipage}[b]{0.55\textwidth}
    \begin{algorithm}[H]
        \small
        \textbf{Input}: LiDAR points $\mathbf{P_{pt}}$, image $\mathbf{I}$, camera parameters $\mT$
        \textbf{Output}: mm-aware \uv features $\mathbf{F}_{pv}$, \bev features $\mathbf{F_{bev}}$, \textit{``mm" denotes multi-modal}.
        \begin{algorithmic}[0] %
        \State $\mathbf{F}_{pt}^{s1} = \textrm{PillarNet-S1}(\mathbf{P}_{pt})$, $\mathbf{F}_{pv}^{s1} = \textrm{ResNet-S1}(\mathbf{I})$ \\
        \Comment{\textit{S1: stage one.}}
        \State $\mathbf{P}_{pt2pv} = \{(u_i, v_i)|i \in P\} = \operatorname{Project}(\mT, \mathbf{P}_{pt})$
        \State $\mathbf{F}_{pt2pv} = \operatorname{Scatter}(idx=\mathbf{P}_{pt2pv}\, src=\mathbf{F}_{pt}^{s1})$ \\
        \Comment{\textit{points $\rightarrow$ pixels.}}
        \State $\mathbf{F}_{pv2pt} = \operatorname{Grid\_Sample}(src=\mathbf{F}^{s1}_{pv}, \,coords=\mathbf{P}_{pt2pv})$ \\
        \Comment{\textit{pixels $\rightarrow$ points.}}
        \State $\mathbf{F}_{pv} = \operatorname{ResNet}(\operatorname{Concat}(\mathbf{F}^{s1}_{pv}, \mathbf{F}_{pt2pv}))$ 
        \State $\mathbf{F}_{bev} = \operatorname{PillarNet}(\operatorname{Concat}(\mathbf{F}^{s1}_{pt}, \mathbf{F}_{pv2pt}))$ \\
        \Comment{\textit{dual-view multi-modal feature extraction.}}
        \end{algorithmic}
        \caption{Bidirectional Feature Fusion (BFF)}
        \label{alg:bff}
    \end{algorithm}
\end{minipage}
\hspace{0.2cm}
\begin{minipage}[b]{0.4\textwidth}
    \includegraphics[width=1\linewidth]{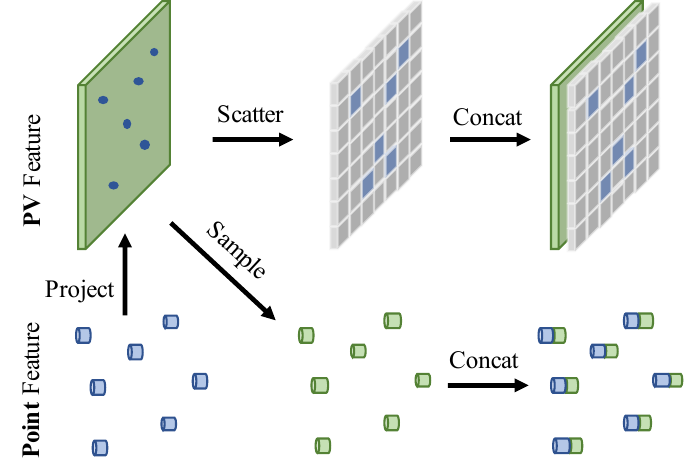}
    \vspace{-5mm}
    \caption{\textbf{Bidirectional Feature Fusion (BFF).} We represent the image feature in green and points in blue.}
    \label{fig:dual_fuse}
\end{minipage}
\end{figure}

Instead of merging different views into one single space~\cite{bai2018deep, luo2022m, liang2022bevfusion, li2022bevformer, liu2023bevfusion,luo2024exploring}, %
we propose to retain features in both \uv and \bev spaces while incorporating multi-modal features for each view.
To achieve this, we employ a dual branch to extract features for each view, using images and points as input, respectively.
Intermediately, we conduct \textit{bidirectional} feature fusion between the symmetric branches to enhance each view with multiple modalities, as shown in~\Cref{fig:dual_fuse} and summarized in~\Cref{alg:bff}.

Concretely, we place points and images in their designated branches.
After obtaining low-level features within each branch, we perform bidirectional feature fusion.
By projecting 3D points $\mathbf{P}_{pt} = \{(x_i, y_i, z_i) | i\in P \} $ onto the \uv plane, we obtain their corresponding 2D coordinates $\mathbf{P}_{pt2pv} = \{(u_i, v_i) | i \in P\}$, where $P$ is the cardinality of the point set. 
\textbf{1)} For \textbf{\textit{points-to-pixels}} fusion, we utilize a $\operatorname{Scatter}$ operation to construct dense point feature grids $\mathbf{F}_{pt2pv}$, (depicted in the upper part of~\Cref{fig:dual_fuse}, with blue cells denoting positions hit by the projected 3D points).
\textbf{2)} For \textbf{\textit{pixels-to-points}} fusion, we employ bilinear interpolation to sample features at 2D positions hit by the projection of 3D points, yielding $\mathbf{F}_{pv2pt}$ (shown in the lower part of~\Cref{fig:dual_fuse}).
The resulting cross-modal features in~\uv and~\bev are concatenated with their respective original modal features.
The fused multi-modal features in each view, \ie,~\uv and~\bev, are then fed into subsequent modules in the corresponding branch, generating $\mathbf{F}_{pv}$ and $\mathbf{F}_{bev}$, respectively. 
Notably, $\mathbf{F}_{pv}$ and $\mathbf{F}_{bev}$ encapsulate multi-modal information represented in distinct spaces.

\subsection{Unified Query Generator}
\label{sec:genquery}
We introduce a unified query generator for end-to-end 3D lane detection. 
To this end, we first generate two distinct lane-aware query sets, termed dual-view queries, 
from the previously obtained multi-modal features, $\mathbf{F}_{pv}$ and $\mathbf{F}_{bev}$.
Then, we present a lane-centric clustering strategy to unify these dual-view queries into a cohesive set of queries.

\begin{wrapfigure}{r}{0.6\linewidth}
    \centering
        \vspace{-3mm}
        \includegraphics[width=1\linewidth]{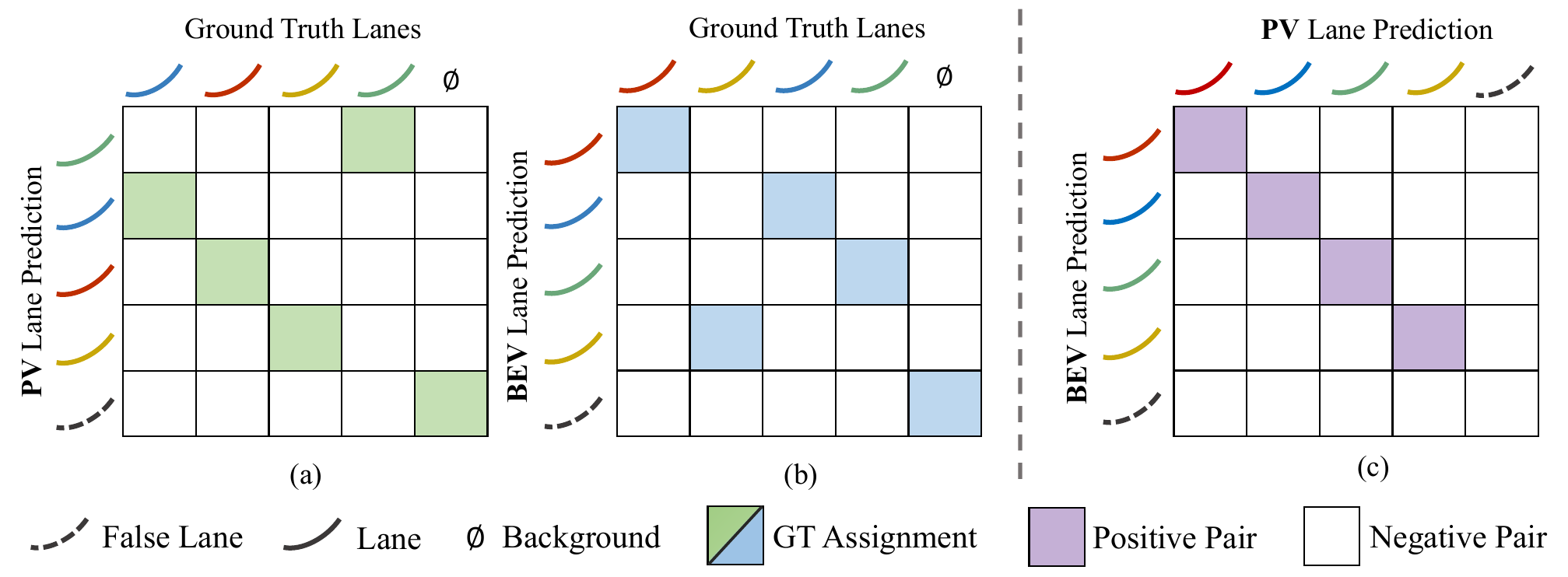}
        \vspace{-6mm}
        \caption{Illustration of one-to-one matching and lane-centric clustering. (a) and (b) show the assignment for \bev and \uv predictions, 
        respectively. (c) depicts the pairing of the clustering, where queries targeting the same lane are treated as a positive pair, otherwise negative.}
        \vspace{2mm}
        \label{fig:cluster_sup}
\end{wrapfigure}

\noindent\textbf{Dual-view Query Generation.}  
To effectively capture semantic and spatial features related to lanes, which are termed as ``lane-aware" knowledge, we utilize an instance activation map (IAM)~\cite{cheng2022sparse}-assisted method to generate lane-aware queries in \uv and \bev spaces. 
Taking \uv branch as an example, we produce a set of IAMs, denoted as $\mathbf{A}_{pv}$, via the following equation:
{
\small
\begin{equation}
\label{eq:iam}
    \mathbf{A}_{pv} = \sigma(\mathcal{F}(\textrm{Concat}(\mathbf{F}_{pv}, \mathbf{S}_{pv}))),
    \nonumber
\end{equation}
\vspace{-0.mm}
}
where $\mathbf{A}_{pv}\in \mathbb{R}^{N \times H_{pv} \times W_{pv}}$, $\mathbf{F}_{pv} \in \mathbb{R}^{C \times H_{pv} \times W_{pv}}$, $N$ denotes query number, $\sigma$ is the sigmoid function, $\textrm{Concat}$ represents concatenation operation, and $\mathbf{S}_{pv}$ comprises two-channel spatial localization features for each pixel~\cite{liu2018intriguing}. 
The lane-aware query $\mathbf{Q}_{pv}$ assisted by IAMs is generated via:
{
\small
\vspace{-2mm}
\begin{equation}
    \mathbf{Q}_{pv} = \mathbf{A}_{pv} \otimes \mathbf{F}_{pv}^\mathsf{T},
    \nonumber
\end{equation}
\vspace{-1mm}
}
where $\mathbf{Q}_{pv} \in \mathbb{R}^{N \times C}$, $\otimes$ denotes the matrix product. Similarly, lane-aware \bev query $\mathbf{Q}_{bev} \in \mathbb{R}^{N \times C}$ is formed using:
{\small
\begin{equation}
    \mathbf{Q}_{bev} = \sigma(\mathcal{F}([\mathbf{F}_{bev}, \mathbf{S}_{bev}])) \otimes \mathbf{F}_{bev}^\mathsf{T}.
    \nonumber
\end{equation}
}
To force the query sets to learn lane-aware features, during training, we employ an auxiliary instance segmentation for each branch on top of the query set.
Labels for the auxiliary segmentation are generated in pairs for these two branches, which are further assigned to predictions using mask-based bipartite matching~\cite{cheng2022sparse}, as illustrated in~\Cref{fig:cluster_sup} (a) and (b).

\noindent\textbf{Dual-view Query Clustering.}
Given dual-view query sets $\mathbf{Q}_{pv}$ and $\mathbf{Q}_{bev}$, we propose employing a \textit{lane-centric} clustering technique to generate a unified query set for end-to-end lane detection.
While $k$Max-DeepLab~\cite{yu2022k} previously used k-means cross-attention to group pixels into distinct clusters, \ie, instance masks, our approach focuses on unifying queries from different views.
Queries from $\mathbf{Q}_{pv}$ and $\mathbf{Q}_{bev}$ targeting the same lane are merged within the same cluster.
Specifically, we initiate lane cluster centers $\mathbf{C} \in \mathbb{R}^{N \times C}$ with $\mathbf{Q}_{pv}$, and assign each query in $\mathbf{Q}_{bev}$ to its nearest cluster center among $\mathbf{C}$.
Notably, cluster centers can be chosen from either $\mathbf{Q}_{pv}$ or $\mathbf{Q}_{bev}$. Empirically, we found that using $\mathbf{Q}_{pv}$ produces better results.
To achieve clustering, we perform attention between $\mathbf{C}$ (query) and $\mathbf{Q}_{bev}$ (key), while applying $\operatorname{argmax}$ along the cluster center (query) dimension~\cite{yu2022k} as follows:
{\small
\begin{equation}
\abovedisplayskip=1.5pt
    \begin{aligned}
    \label{eq:kmeans_update}
    \mathbf{A} = \operatornamewithlimits{argmax}_{N}(\mathbf{C} \times \mathbf{Q}_{bev}^{\mathsf{T}}),
    \quad
    \hat{\mathbf{C}} = \mathbf{A} \cdot \mathbf{Q}_{bev} + \mathbf{C},
    \end{aligned}
    \nonumber
\belowdisplayskip=-0.7pt
\end{equation}
}
where $\hat{\mathbf{C}} \in \mathbb{R}^{N \times C}$ refers to updated centers unifying queries from dual views. 
In practise, we use $\operatorname{gumbel-softmax}$~\cite{jang2016categorical,liang2023clustseg} to substitute $\operatorname{argmax}$. 

Considering the variation and slenderness of lanes, we employ a refined point query scheme~\cite{luo2023latr} to enhance lane detection.
Instead of using a single query for each lane, multiple-point queries are employed for more precise capture~\cite{luo2023latr,liao2022maptr,zhang2021direct,liu2023group}.
Consequently, 
in the first layer, we construct point-based queries $\mathbf{Q} \in \mathbb{R}^{N \times M \times C}$ with $\mathbf{Q} = \hat{\mathbf{C}} \oplus \mathbf{E}_{points}$, 
where $\oplus$ denotes broadcast sum, $\mathbf{E}_{points} \in \mathbb{R}^{M \times C}$ is the learnable point embedding, and in the subsequent layer, we update $\mathbf{Q}$ by $\mathbf{Q} = \hat{\mathbf{C}} \oplus \mathbf{Q}$.

\noindent\textbf{Supervision on Query Clustering.}
Given the critical importance of deep supervision for the clustering~\cite{yu2022k}, we leverage the InfoNCE loss~\cite{oord2018representation, radford2021learning}, to supervise the query clustering in a lane-centric manner, as illustrated in~\Cref{fig:cluster_sup} (c) and formulated as~\Cref{eq:infoNCE},
where $\tau$ is a temperature hyper-parameter~\cite{wu2018unsupervised}, $\bm{q}$ denotes one query, $\bm{k}^{+}$ indicates the positive sample \wrt $\bm{q}$, and $\mathcal{N}$ denotes the collection of all negative samples from the different query set relative to the one containing $\bm{q}$.
Notably, queries assigned to the background do not incur penalties in the clustering learning process.
With this supervision, queries from different views are grouped together when matched to the same ground truth lane. Consequently, lane-aware knowledge residing in two view spaces is synergized into the unified query.
\vspace{-2mm}
{\small
\begin{equation}
\label{eq:infoNCE}
\!\!\!\!\mathcal{L}_{\text{NCE}}\!=-{\!}\log\frac{\exp(\bm{q}\!\cdot\!\bm{k}^{+\!\!}/\tau)}{\exp(\bm{q}\!\cdot\!\bm{k}^{+\!\!}/\tau)
+\!\sum\nolimits_{\bm{k}^{-\!}\in\mathcal{N}}\exp(\bm{q}\!\cdot\!\bm{k}^{-\!\!}/\tau)},\!\!
\end{equation}
}

\subsection{3D Dual-view Deformable Attention}
\label{sec:dualattn}
\begin{figure}[h]
\vspace{-3mm}
\centering
    \begin{minipage}[b]{0.5\textwidth}
        \begin{algorithm}[H]
        \small
        \caption{3D DV Deformable Attention} 
        \label{alg:ddda}
        \textbf{Input}: unified query set $\mathbf{Q}$, \uv features $\mathbf{F}_{pv}$, \bev features $\mathbf{F}_{bev}$, camera parameters $\textrm{T}$.\\
        \textbf{Output}: updated unified query $\mathbf{Q}$.
        \begin{algorithmic}[0]
        \State $\mathbf{Ref}_{3d}$ = $\textrm{MLP}_{1}(\mathbf{Q})$ \\
        \Comment{\textit{3D reference points.}}
        \State $\mathbf{\Delta Ref}_{3d} = \textrm{MLP}_{2}(\mathbf{Q})$ %
        \State $\mathbf{S}_{3d} = \{(x_i, y_i, z_i) | i \in N\} = \mathbf{\Delta Ref}_{3d} + \mathbf{Ref}_{3d} $
        \\
        \Comment{\textit{deformed 3D positions.}}
        \State $\mathbf{D}_{pv} = \operatorname{DeformAttn}(\operatorname{Project_{pv}(\mathbf{S}_{3d}, T)}, \mathbf{F}_{pv})$
        \\
        \Comment{\textit{project 3D deformed points to PV.}}
        \State $\mathbf{D}_{bev} = \operatorname{DeformAttn}(\operatorname{Project}_{bev}(\mathbf{S}_{3d}), \mathbf{F}_{bev})$
        \State $\mathbf{Q} = \textrm{SE}(\mathbf{D}_{pv} , \mathbf{D}_{bev})$ 
        \end{algorithmic}
        \end{algorithm}
    \end{minipage}
\hspace{5mm}
    \begin{minipage}[b]{0.45\textwidth}
        \vspace{2mm}
        \includegraphics[width=0.92\linewidth]{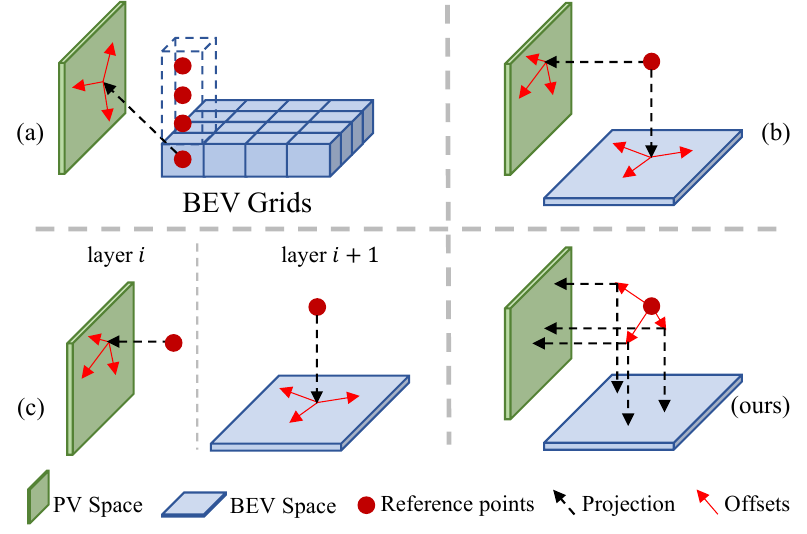}
        \vspace{-2mm}
        \caption{Illustration comparing 3D dual-view deformable attention with other approaches.}
        \label{fig:3d_deform}
    \end{minipage}
    \vspace{-4mm}
\end{figure}

Apart from informative query generation, feature aggregation plays a crucial role in~\modelname.
Instead of projecting points from densely sampled grids~\cite{chen2022persformer} or their lifted pillars~\cite{li2022bevformer} onto the \uv plane for feature sampling, as shown in~\Cref{fig:3d_deform} (a), 
we adopt sparse queries to sample features from different views.
Moreover,
our approach distinguishes itself from several existing sparse query methods, as depicted in~\Cref{fig:3d_deform} (b) and (c).
For instance,
DeepInteration~\cite{yang2022deepinteraction} (\Cref{fig:3d_deform} (b)) employs a sequential method to sample \uv and \bev features,
while FUTR3D~\cite{chen2023futr3d} (\Cref{fig:3d_deform} (c)) projects 3D points into different spaces, sampling features individually for each space.

In contrast, as outlined in~\Cref{alg:ddda}, we leverage the inherent properties of 3D space by predicting both 3D reference points and their 3D offsets~\cite{luo2022detr4d} using queries, forming 3D deformed points. 
These 3D deformed points are then projected into each space, establishing a \textit{consistent} feature sampling strategy across spaces, as depicted in~\Cref{fig:3d_deform}.
Consequently, features corresponding to the same 3D points from different views are effectively sampled and integrated into the query.

\subsection{Prediction and Loss}
\noindent \textbf{Auxiliary Tasks.} During training, we incorporate two auxiliary tasks: 1) 2D instance segmentation~\cite{luo2023latr,cheng2022sparse} loss $\mathcal{L}_{seg}$ for both \uv and \bev branches, aiding in extracting discriminative lane features in each view;
2) Depth estimation for the \uv branch, which guides effective 3D structure-aware feature extraction of $\mathbf{F}_{pv}$. Depth labels are generated from LiDAR points, and the loss $\mathcal{L}_{depth}$ is calculated following BEVDepth~\cite{li2022bevdepth}.

\noindent \textbf{3D Lane Prediction and Loss.}
As we adopt point-based queries $\mathbf{Q}\in\mathbb{R}^{(N \times M) \times C}$, each query naturally corresponds to a 3D point, and every group of $M$ points constructs a complete 3D lane.
Thus, we predict x, z, and visibility for each point query on the predefined y coordinates~\cite{chen2022persformer,luo2023latr} and a classification probability for each lane. Overall, the total loss is:
\begin{equation}
    \begin{aligned}
        \mathcal{L}_{lane} &= w_{x}\mathcal{L}_{x} + w_{z}\mathcal{L}_{z} + w_{v}\mathcal{L}_{v} + w_{c}\mathcal{L}_{c}, \\
        \mathcal{L}_{aux} &= w_{seg}\mathcal{L}_{seg} + w_{depth}\mathcal{L}_{depth}, \\
        \mathcal{L}_{total} &= \mathcal{L}_{lane} + \mathcal{L}_{aux}.
    \end{aligned}
    \nonumber
\end{equation}
where $w_*$ denotes different loss weights. We adopt the L1 loss $\mathcal{L}_{x}$ and $\mathcal{L}_{z}$ to learn the x, z positions, focal loss~\cite{lin2017focal} $\mathcal{L}_{c}$ to learn the lane category, and BCELoss $\mathcal{L}_{v}$ to learn visibility. 

\section{Experiments}

\subsection{Datasets}
We evaluate our method on OpenLane~\cite{chen2022persformer}, the \textit{sole} public 3D lane dataset featuring multi-modal sources,
OpenLane is a large-scale dataset built on Waymo Open Dataset~\cite{sun2020scalability}, comprising 200K frames and 880K lanes across six driving scenarios and 14 lane categories.
The LiDAR data, collected using 64-beam LiDARs, is sampled at 10Hz.
This extensive dataset provides a solid foundation for evaluating 3D lane algorithms comprehensively.

\begin{wrapfigure}{r}{0.47\linewidth}
    \centering
        \vspace{-4mm}
        \includegraphics[width=1\linewidth]{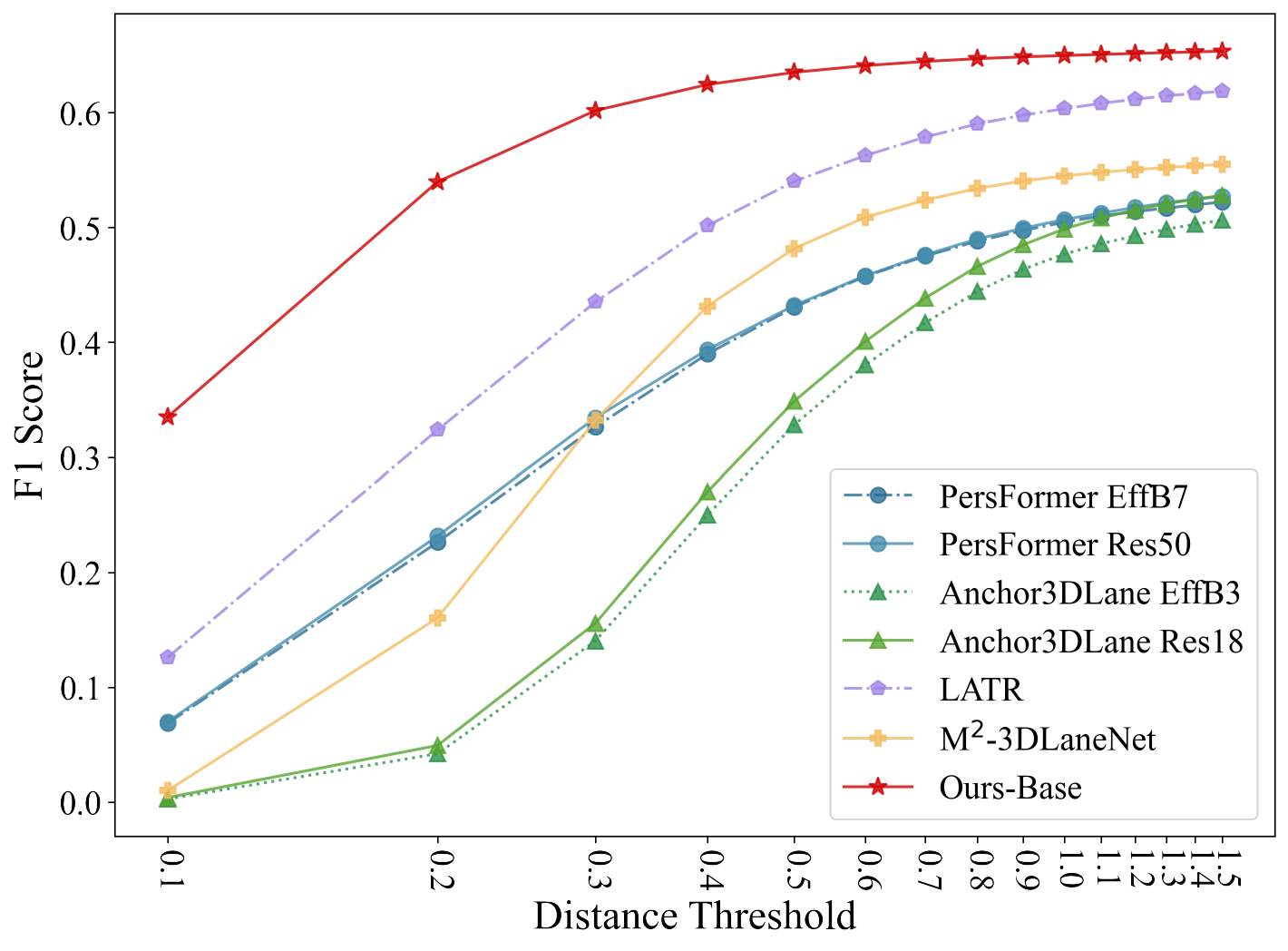}
        \vspace{-6mm}
        \caption{\textbf{F1 score vs. Distance Threshold.} Our method consistently achieves superior results under more stringent criteria.}
        \label{fig:f1_vs_distance}
        \vspace{-10mm}
\end{wrapfigure}

\subsection{Metrics}
We adopt the evaluation metrics established by OpenLane~\cite{chen2022persformer}, framing 3D lane detection evaluation as a matching problem based on the edit distance between predictions and ground truth.
Successful matching results in computed metrics, including F-Score, category accuracy, and error in X/Z-axes.
A successful match for each predicted 3D lane is defined when at least 75\% of its points have a distance to the ground truth below the predefined threshold $D_{thre}$.

\subsection{Implementation Details}

\noindent\textbf{Models.} In the base version of~\modelname, we employ ResNet34~\cite{he2016deep} and PillarNet34~\cite{shi2022pillarnet} as the backbones for our camera and LiDAR branches, respectively. For the lite version, we utilize ResNet18 and PillarNet18. The base version features two decoder layers, while the lite version employs a single decoder layer.
Following LATR~\cite{luo2023latr}, we set the number of lane queries to 40, and we employ deformable attention with 4 heads, 8 sample points, and 256 embedding dimensions.

\noindent\textbf{Training.} We use the Adam optimizer~\cite{kingma2014adam} with a weight decay of 0.01. The learning rate is set to 2e-4, and our models undergo training for 24 epochs with a batch size of 32. We employ the cosine annealing scheduler~\cite{loshchilov2016sgdr} with $T_{max}=8$.
Our input images are of resolution 720$\times$960, and we adopt a voxel size of (0.2m, 0.4m) for the X and Y axes. 

\subsection{Main Results}

It's important to note that the existing metrics use a rather \textit{lenient} distance threshold of $D_{thre}$=\textbf{1.5m}. However, this value, although commonly used for assessment purposes, may be considered overly permissive in the context of ensuring safety in AD. 
Following M$^2$-3DLaneNet~\cite{luo2022m}, we extend our evaluation to include a \textit{more stringent} threshold, $D_{thre}$=\textbf{0.5m}.
Further, we illustrate the relationship between the F1 score performance and different distance thresholds for various models, as shown in~\Cref{fig:f1_vs_distance}. 
Notably, our method consistently achieves superior results, even when evaluated under a \textit{much more stringent} criterion of $D_{thre}$=\textbf{0.1m}. In contrast, other approaches experience a noticeable decline in performance as the distance threshold decreases. 
These findings confirm the robustness of our method across varying distance thresholds, particularly highlighting its advantage in precise localization.

We present the main results in~\Cref{tab:main_results}, obtained from experiments conducted on the OpenLane-1K dataset.
The evaluation uses both $D_{thre}$=1.5m and $D_{thre}$=0.5m criteria, allowing for a comprehensive and insightful comparison.
It is evident that~\modelname consistently outperforms previous state-of-the-art (SoTA) methods across all metrics. Notably, when applying a more strict 0.5m threshold,~\modelname demonstrates a substantial {\textbf{11.2\%}} improvement in the F1 score.
Notably, our method excels in localization accuracy, leading to significant performance improvements. 
Specifically, our method achieves remarkable reductions in localization errors: {52\%/50\%} for X near/far and 
{61\%/52\%} for Z near/far. Due to space limitations, results in various scenarios and studies about robustness concerning calibration noise are included in the Appendix.

\noindent \textbf{Effect of Multiple Modalities.}
To explore the impact of individual modalities, we conduct experiments using single modalities, as outlined in the ``Image-Branch" and ``LiDAR-Branch" rows of~\Cref{tab:main_results}.
The results illustrate that~\modelname significantly enhances performance compared to using images alone or relying solely on LiDAR data.
Notably, our method significantly surpasses configurations that simply equip LATR with LiDAR input across all metrics, underscoring the substantial improvements achieved by~\modelname in leveraging information from both modalities.
Moreover, to evaluate the effect of \textbf{dual-view}, we conduct experiments using single-modality input but transforming features extracted from the backbone into another view, yielding single-modal dual-view features. Then, our dual-view decoder is applied, and the results are detailed in the Appendix.
Additionally, we conduct experiments using our ``Image-Branch" on the Apollo~\cite{guo2020gen} dataset, which exclusively contains image data. The results are provided in the Appendix.

\begin{figure*}[h]\centering
    \vspace{-2mm}
   \includegraphics[width=1.0\linewidth]{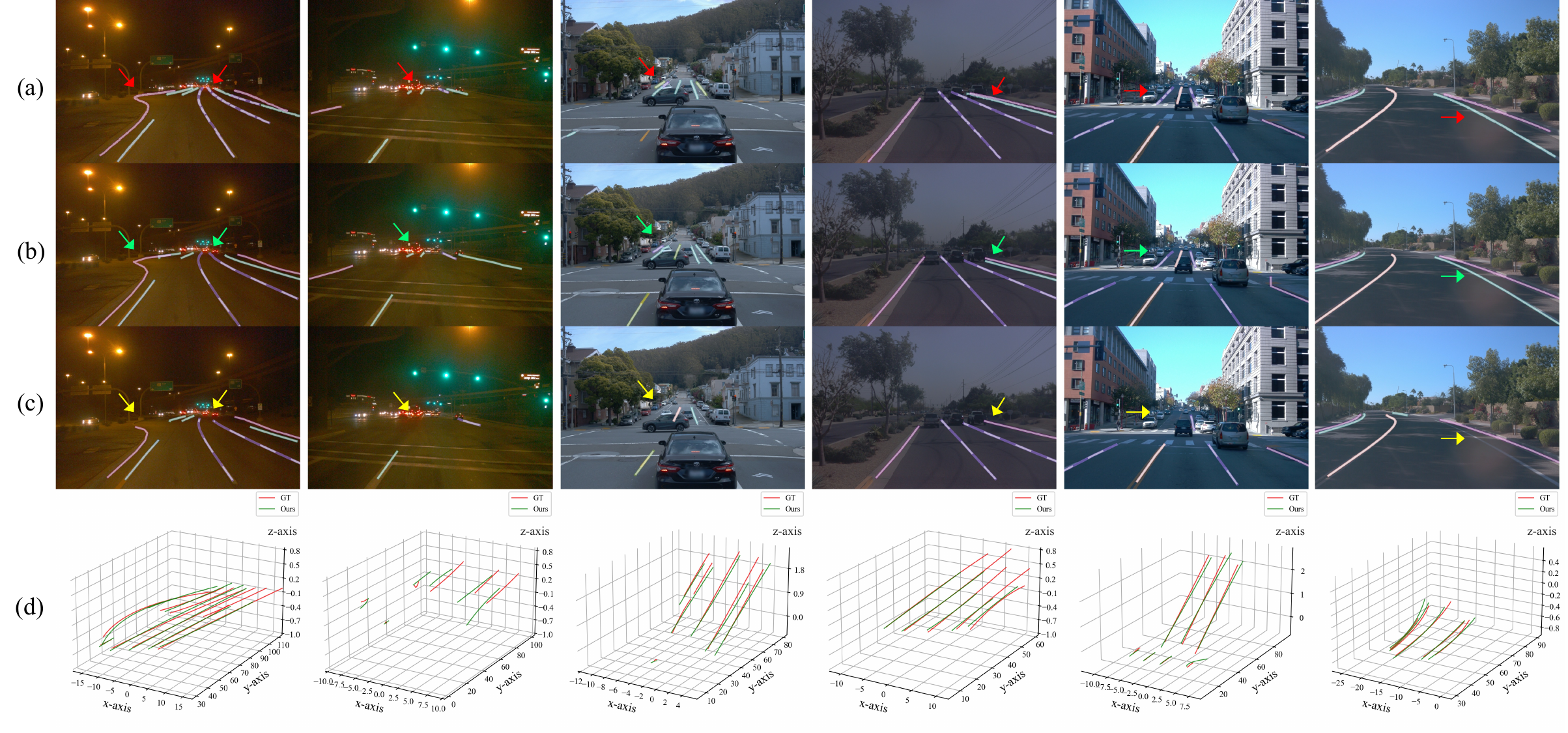}
   \vspace{-6mm}
    \caption{\textbf{Qualitative Results.} We present the projection of 3D lanes from \textcolor{gt}{ground truth}, predictions of \textcolor{Green}{\modelname} and the SOTA method \textcolor{prev_sota}{LART}~\cite{luo2023latr} in rows \textcolor{gt}{(a)}, \textcolor{Green}{(b)}, \textcolor{prev_sota}{(c)}, respectively. Row (d) depicts the comparison between \textcolor{gt}{ground truth} (red) and \textcolor{Green}{ours} (green) in 3D space. We highlight the differences with colored arrows. Best viewed in color and zoom in for details.}
\label{fig:qua_results}
\vspace{-3mm}
\end{figure*}

\noindent\textbf{Qualitative Results.} We present a qualitative comparison between~\modelname and LATR~\cite{luo2023latr} in~\Cref{fig:qua_results}, demonstrating that our method achieves more robust and accurate predictions across various scenarios. More visualization results are included in the Appendix.

\begin{table*}[tp] 
\centering
    \footnotesize
    \caption{Comprehensive 3D Lane evaluation comparison on OpenLane with variable metrics. $\dagger$ denotes the results obtained using their provided models. ``Image-Branch" and ``LiDAR-Branch" refer to our image and LiDAR branches, respectively. ``LATR + LiDAR" denotes the model that combines the SOTA method LATR with LiDAR input, projecting all points into the image space and using them as additional features in the network. }
    \label{tab:main_results}
    \vspace{2mm}
    \resizebox{1.0\textwidth}{!}{
    \begin{tabular}{c|c|c|c|c|c|*{2}{c}|*{2}{c}}
    \Xhline{1pt}
    \multirow{2}{*}{\textbf{\textit{Dist.}}} & \multirow{2}{*}{\textbf{Methods}} & \multirow{2}{*}{\textbf{Backbone}} 
    & \multirow{2}{*}{\textbf{Modality}} 
    & \multirow{2}{*}{\textbf{F1}\textcolor{impblue}{\pmb{ $\uparrow$}}}
    & \multirow{2}{*}{\textbf{\textit{Acc.}}\textcolor{impblue}{\pmb{ $\uparrow$}}}
    & \multicolumn{2}{c}{\textbf{X error (m)}\textcolor{impblue}{\pmb{ $\downarrow$}}}
    & \multicolumn{2}{c}{\textbf{Z error (m)}\textcolor{impblue}{\pmb{ $\downarrow$}}} \\
    \cline{7-10}
    & & & & &  & \near & \far & \near & \far \\
    \Xhline{0.5pt}
    \addlinespace[1.2pt]
    \multirow{12}{*}{{\rotatebox{90}{\small{\textbf{1.5 m}}}}} 
    & \multicolumn{1}{l|}{3DLaneNet~\cite{garnett20193d}} & VGG-16 & C & 44.1 & - & 0.593 & 0.494 & 0.140 & 0.195 \\
    & \multicolumn{1}{l|}{GenLaneNet~\cite{guo2020gen}} & ERFNet & C & 32.3 & - & 0.591 & 0.684 & 0.411 & 0.521 \\
    & \multicolumn{1}{l|}{PersFormer~\cite{chen2022persformer}} & EffNet-B7 & C & 50.5 & 89.5 & 0.319 &0.325 & 0.112 &0.141 \\
    & \multicolumn{1}{l|}{Anchor3DLane~\cite{huang2023anchor3dlane}$^\dagger$} & EffNet-B3 & C & 52.8 & 89.6 & 0.408 & 0.349 & 0.186 & 0.143 \\
    & \multicolumn{1}{l|}{M$^2$-3DLaneNet~\cite{luo2022m}} & EffNet-B7 & C+L & 55.5 & 88.2 & 0.283 & \underline{0.256} & 0.078 & 0.106 \\
    & \multicolumn{1}{l|}{Anchor3DLane~\cite{huang2023anchor3dlane}$^\dagger$} & ResNet-18 & C & 50.7 & 89.3 & 0.422 & 0.349 & 0.188 & 0.146 \\
    & \multicolumn{1}{l|}{PersFormer~\cite{chen2022persformer}} & ResNet-50 & C & 52.7 & 88.4 & 0.307 & 0.319 & 0.083 & 0.117 \\
    & \multicolumn{1}{l|}{LATR~\cite{luo2023latr}} & ResNet-50 & C & \underline{61.9} & \underline{92.0} & \underline{0.219} & 0.259 & \underline{0.075} & \underline{0.104} \\
    \cline{2-10}
    \addlinespace[1.2pt]
    & \multicolumn{1}{l|}{\cellcolor{mygray}{\modelname-{Tiny} (Ours)}} & \cellcolor{mygray}{ResNet-18} & \cellcolor{mygray}{C+L} & \cellcolor{mygray}{63.4} & \cellcolor{mygray}{91.6} & \cellcolor{mygray}{0.137} & \cellcolor{mygray}{0.159} & \cellcolor{mygray}{0.034} & \cellcolor{mygray}{0.063} \\
    & \multicolumn{1}{l|}{\cellcolor{mygray}{\modelname-{Base} (Ours)}} & \cellcolor{mygray}{ResNet-34} & \cellcolor{mygray}{C+L} & \cellcolor{mygray}{{65.4}} & \cellcolor{mygray}{{92.4}} & \cellcolor{mygray}{{0.118}} & \cellcolor{mygray}{{0.131}} & \cellcolor{mygray}{{0.032}} & \cellcolor{mygray}{{0.053}} \\
    & \multicolumn{1}{l|}{\cellcolor{mygray}{{\modelname-Large (Ours)}}} & \cellcolor{mygray}{{ResNet-50}} & \cellcolor{mygray}{{C+L}} & \cellcolor{mygray}{{\textbf{66.8}}} & \cellcolor{mygray}{{\textbf{93.3}}} & \cellcolor{mygray}{{\textbf{0.115}}} & \cellcolor{mygray}{{\textbf{0.134}}} & \cellcolor{mygray}{{\textbf{0.029}}} & \cellcolor{mygray}{{\textbf{0.049}}} \\
    \cline{2-10}
    \addlinespace[1.2pt]
    & \emph{Improvement} & - & - 
    & \textcolor{impblue}{\textit{$\uparrow$4.9}}
    & \textcolor{impblue}{\textit{$\uparrow$1.3}}
    & \textcolor{impblue}{\textit{$\downarrow$0.104}} 
    & \textcolor{impblue}{\textit{$\downarrow$0.122}}
    & \textcolor{impblue}{\textit{$\downarrow$0.046}} 
    & \textcolor{impblue}{\textit{$\downarrow$0.055}} \\
    \Xhline{0.6pt}
    \addlinespace[1.2pt]
    \multirow{13}{*}{{\rotatebox{90}{\small{\textbf{0.5 m}}}}}
    & \multicolumn{1}{l|}{PersFormer~\cite{chen2022persformer}} & EffNet-B7 & C & 36.5 & 87.8 & 0.343 & 0.263 & 0.161 & 0.115 \\
    & \multicolumn{1}{l|}{Anchor3DLane~\cite{huang2023anchor3dlane}$^\dagger$} & EffNet-B3 & C & 34.9 & 88.5 & 0.344 & 0.264 & 0.181 & 0.134 \\
    & \multicolumn{1}{l|}{M$^2$-3DLaneNet~\cite{luo2022m}} & EffNet-B7 & C+L & 48.2 & 88.1 & 0.217 & 0.203 & 0.076 & 0.103  \\
    & \multicolumn{1}{l|}{Anchor3DLane~\cite{huang2023anchor3dlane}$^\dagger$} & ResNet-18 & C & 32.8 & 87.9 & 0.350 & 0.266 & 0.183 & 0.137 \\
    & \multicolumn{1}{l|}{PersFormer~\cite{chen2022persformer}} & ResNet-50 & C & 43.2 & 87.8 & 0.229 & 0.245 & 0.078 & 0.106  \\
    & \multicolumn{1}{l|}{LATR~\cite{luo2023latr}} & ResNet-50 & C & \underline{54.0} & \underline{91.7} & \underline{0.171} & \underline{0.201} & \underline{0.072} & \underline{0.099} \\
    \cline{2-10}
    \addlinespace[1.2pt]
     & \multicolumn{1}{l|}{{LATR + LiDAR}} & {ResNet-50} & {C+L} & {57.4} & {92.1} & {0.167} & {0.185} & {0.071} & {0.088} \\
    & \multicolumn{1}{l|}{{Image-Branch (Ours)}} & {ResNet-34} & {C} & {52.9} & {90.3} & {0.173} & {0.212} & {0.069} & {0.098} \\
    & \multicolumn{1}{l|}{LiDAR-Branch (Ours)} & PillarN-34 & L & 54.1 & 84.4 & 0.282 & 0.191 & 0.096 & 0.124 \\
    & \multicolumn{1}{l|}{\cellcolor{mygray}{\modelname-Tiny} (Ours)} & \cellcolor{mygray}{ResNet-18} & \cellcolor{mygray}{C+L} & \cellcolor{mygray}{60.9} & \cellcolor{mygray}{91.8} & \cellcolor{mygray}{0.097} & \cellcolor{mygray}{0.124} & \cellcolor{mygray}{0.033} & \cellcolor{mygray}{0.062} \\
    & \multicolumn{1}{l|}{\cellcolor{mygray}{\cellcolor{mygray}{\modelname-Base} (Ours)}} & \cellcolor{mygray}{ResNet-34} & \cellcolor{mygray}{C+L} & \cellcolor{mygray}{{63.5}} & \cellcolor{mygray}{{92.4}} & \cellcolor{mygray}{{0.090}} & \cellcolor{mygray}{{0.102}} & \cellcolor{mygray}{{0.031}} & \cellcolor{mygray}{{0.053}} \\
    & \multicolumn{1}{l|}{\cellcolor{mygray}{\modelname-Large (Ours)}} & \cellcolor{mygray}{ResNet-50} & \cellcolor{mygray}{C+L} & \cellcolor{mygray}{\textbf{65.2}} & \cellcolor{mygray}{\textbf{93.4}} & \cellcolor{mygray}{\textbf{0.082}} & \cellcolor{mygray}{\textbf{0.101}} & \cellcolor{mygray}{\textbf{0.028}} & \cellcolor{mygray}{\textbf{0.048}} \\
    \cline{2-10}
    \addlinespace[1.2pt]
    & \emph{Improvement} & - & - 
    & \textcolor{impblue}{\textit{$\uparrow$11.2}}
    & \textcolor{impblue}{\textit{$\uparrow$1.7}}
    & \textcolor{impblue}{\textit{$\downarrow$0.089}}
    & \textcolor{impblue}{\textit{$\downarrow$0.100}}
    & \textcolor{impblue}{\textit{$\downarrow$0.044}}
    & \textcolor{impblue}{\textit{$\downarrow$0.051}} \\
    \Xhline{1pt}
    \end{tabular}
    }
    \vspace{-5mm}
\end{table*}

\subsection{Ablation Studies}

We conduct all ablation studies on OpenLane-300 following established practices~\cite{chen2022persformer,luo2023latr,huang2023anchor3dlane}, while adopting a \textbf{0.5m} threshold $D_{thre}$ for evaluation. 

\noindent\textbf{Effect of Bidirectional Feature Fusion.} 
The corresponding experiments are included in the Appendix due to space limitations. We kindly direct the readers to refer to the Appendix for details. The results confirm the effectiveness of the proposed bidirectional feature fusion approach.

\noindent\textbf{Effect of Unified Query.}
We study the effect of our unified query generation strategy in~\Cref{tab:query_cluster}, where
``Random" means random initialization using \texttt{nn.Embedding}, ``$\mathbf{Q}_{pv}$" denotes using only \uv queries, and ``$\mathbf{Q}_{bev}$" refers to using only \bev queries.
Replacing our unified queries with randomly initialized ones~\cite{carion2020end,zhu2020deformable,li2022bevformer} results in a decrease of 1.0 in the F1 score compared to our approach. Interestingly, employing a single space instance-aware query yields even lower F1 scores of 69.6\%/69.1\% for \uv/\bev, respectively, than random initialization. 
This underscores the inadequacy of a single-space lane-aware query in capturing complex 3D lane features comprehensively existing in both \uv and \bev spaces.
However, our dual-view strategy, generating lane-aware queries \wrt both views, improves overall performance to 70.7, achieving the best result. 
This demonstrates that our method effectively integrates the strengths of features from two spaces, forming a cohesive query set.

\noindent\textbf{Effect of 3D Dual-view Deformable Attention.}
To evaluate the efficacy of our proposed Dual-view Deformable Attention, we conduct ablation studies in~\Cref{tab:3d_deform}, where ``\uv space" and ``\bev space" mean using single space in the decoder. ``DeepInteration"~\cite{yang2022deepinteraction} denotes sequential fusion of features from different spaces, and ``FUTR3D"~\cite{chen2023futr3d} refer to a modality-agnostic approach where sampling locations differ across views. We compare \modelname against alternative approaches, including the single-view fused method, as well as methods proposed in DeepInteration and FUTR3D, as described in~\Cref{sec:dualattn}.
The results underscore the significance of our approach. 
In detail, sampling only \uv space features leads to a notable drop (70.7$\rightarrow$63.6) in performance, showing the importance of BEV space due to its advantages in localization. 
Besides, our method outperforms the sequential approach of DeepInteration with a substantial 2.0 gain in F1 score. 
Furthermore, compared to the modality-agnostic approach proposed in FUTR3D, our method achieves a 0.5 improvement, emphasizing the importance of consistent sampling locations in deformable attention across different spaces.

\begin{table}[h]
\vspace{-5mm}
\begin{minipage}[b]{0.42\textwidth}
\vspace{-4mm}
\caption{Effect of unified query.}
\vspace{1mm}
    \centering
    \label{tab:query_cluster}
    \scalebox{0.8}{
    \begin{tabular}{cc|c|c|c}
        \toprule[1.2pt]
        \multicolumn{2}{c|}{\multirow{2}{*}{Methods}}  & \multirow{2}{*}{F1} & {X error (m)} & {Z error (m)} \\
        & & & \multicolumn{1}{c|}{\makebox[0pt][c]
        {\hspace{-1.7em}\textit{near}\hspace{1em}}$\vert$\makebox[0pt][c]{\hspace{3.2em}\textit{far}\hspace{0.5em}}} & \multicolumn{1}{c}{\makebox[0pt][c]{\hspace{-1.7em}\textit{near}\hspace{1em}}$\vert$\makebox[0pt][c]{\hspace{3.2em}\textit{far}\hspace{0.5em}}}\\
        \hline
        \addlinespace[1.2pt]
        \multicolumn{2}{c|}{Random} & 69.7 & 0.123 $\vert$ 0.151 & 0.059 $\vert$ 0.081 \\ %
        \multicolumn{2}{c|}{$\mathbf{Q}_{pv}$} &  69.6 & 0.124 $\vert$ 0.155 & 0.059 $\vert$ 0.079 \\
        \multicolumn{2}{c|}{$\mathbf{Q}_{bev}$} & 69.1 & 0.122 $\vert$ 0.145 & 0.058 $\vert$ 0.077 \\
        \multicolumn{2}{c|}{Ours} & 70.7 & 0.123 $\vert$ 0.146 & 0.058 $\vert$ 0.078 \\
        \toprule[1.2pt]
    \end{tabular}
    }
\end{minipage}
\hspace{2mm}
\begin{minipage}[b]{0.57\textwidth}
    \caption{Effect of 3D dual-view deformable attention.}
    \vspace{2mm}
    \centering
    \label{tab:3d_deform}
    \scalebox{0.8}{
    \begin{tabular}{c|c|c|c}
        \toprule[1.2pt]
        \multirow{2}{*}{Methods}  & \multirow{2}{*}{F1} & {X error (m)} & {Z error (m)} \\
        & & \multicolumn{1}{c|}{\makebox[0pt][c]{\hspace{-1.7em}\textit{near}\hspace{1em}}$\vert$\makebox[0pt][c]{\hspace{3.2em}\textit{far}\hspace{0.5em}}} & \multicolumn{1}{c}{\makebox[0pt][c]{\hspace{-1.7em}\textit{near}\hspace{1em}}$\vert$\makebox[0pt][c]{\hspace{3.2em}\textit{far}\hspace{0.5em}}}\\
        \hline
        \addlinespace[1.2pt]
         \multicolumn{1}{l|}{\uv space} &  63.6 & 0.150 $\vert$~0.202 & 0.060 $\vert$ 0.081 \\
         \multicolumn{1}{l|}{\bev space} & 68.5 & 0.127 $\vert$~0.151 & 0.064 $\vert$~0.087 \\
          \multicolumn{1}{l|}{DeepInteration} & 68.7 & 0.126 $\vert$ 0.157 & 0.059 $\vert$ 0.081 \\
           \multicolumn{1}{l|}{FUTR3D} & 70.2 & 0.118 $\vert$ 0.145 & 0.057 $\vert$ 0.077 \\
         \multicolumn{1}{l|}{Ours} & 70.7 & 0.123 $\vert$ 0.146 & 0.058 $\vert$ 0.078 \\
    \toprule[1.2pt]
    \end{tabular}
    }
\end{minipage}
\vspace{-7mm}
\end{table}

\section{Conclusion}
In this work, we introduce~\modelname, a novel end-to-end multi-modal 3D lane detection framework that leverages the strengths of both \uv and \bev spaces.
To this end, we propose three novel modules that effectively utilize dual-view representation on different levels, consistently enhancing performance.
Extensive experiments substantiate the outstanding advancements achieved by~\modelname, establishing a new state of the art on OpenLane.

\section*{Acknowledgments}
This work was supported by NSFC with Grant No. 62293482, by the Basic Research Project No. HZQB-KCZYZ-2021067 of Hetao Shenzhen HK S$\&$T Cooperation Zone, by Shenzhen General Program No. JCYJ20220530143600001, by Shenzhen-Hong Kong Joint Funding No. SGDX20211123112401002, by the National Key R$\&$D Program of China with grant No. 2018YFB1800800, by the Shenzhen Outstanding Talents Training Fund 202002, by Guangdong Research Project No. 2017ZT07X152 and No. 2019CX01X104, by the Guangdong Provincial Key Laboratory of Future Networks of Intelligence (Grant No. 2022B1212010001), by the Guangdong Provincial Key Laboratory of Big Data Computing, The Chinese University of Hong Kong, Shenzhen, by the NSFC 61931024$\&$12326610, by the Shenzhen Key Laboratory of Big Data and Artificial Intelligence (Grant No. ZDSYS201707251409055), and the Key Area R$\&$D Program of Guangdong Province with grant No. 2018B03033800, by Tencent$\&$Huawei Open Fund.

\bibliography{iclr2024_conference}

\begin{thebibliography}{78}
\providecommand{\natexlab}[1]{#1}
\providecommand{\url}[1]{\texttt{#1}}
\expandafter\ifx\csname urlstyle\endcsname\relax
  \providecommand{\doi}[1]{doi: #1}\else
  \providecommand{\doi}{doi: \begingroup \urlstyle{rm}\Url}\fi

\bibitem[Ai et~al.(2023)Ai, Ding, Zhao, and Zhong]{ai2023ws}
Jianyong Ai, Wenbo Ding, Jiuhua Zhao, and Jiachen Zhong.
\newblock Ws-3d-lane: Weakly supervised 3d lane detection with 2d lane labels.
\newblock In \emph{2023 IEEE International Conference on Robotics and Automation (ICRA)}, pp.\  5595--5601. IEEE, 2023.

\bibitem[Bai et~al.(2018)Bai, Mattyus, Homayounfar, Wang, Lakshmikanth, and Urtasun]{bai2018deep}
Min Bai, Gellert Mattyus, Namdar Homayounfar, Shenlong Wang, Shrinidhi~Kowshika Lakshmikanth, and Raquel Urtasun.
\newblock Deep multi-sensor lane detection.
\newblock In \emph{2018 IEEE/RSJ International Conference on Intelligent Robots and Systems (IROS)}, pp.\  3102--3109. IEEE, 2018.

\bibitem[Bai et~al.(2022{\natexlab{a}})Bai, Hu, Zhu, Huang, Chen, Fu, and Tai]{bai2022transfusion}
Xuyang Bai, Zeyu Hu, Xinge Zhu, Qingqiu Huang, Yilun Chen, Hongbo Fu, and Chiew-Lan Tai.
\newblock Transfusion: Robust lidar-camera fusion for 3d object detection with transformers.
\newblock In \emph{Proceedings of the IEEE/CVF conference on computer vision and pattern recognition}, pp.\  1090--1099, 2022{\natexlab{a}}.

\bibitem[Bai et~al.(2022{\natexlab{b}})Bai, Chen, Fu, Peng, Liang, and Cheng]{bai2022curveformer}
Yifeng Bai, Zhirong Chen, Zhangjie Fu, Lang Peng, Pengpeng Liang, and Erkang Cheng.
\newblock Curveformer: 3d lane detection by curve propagation with curve queries and attention.
\newblock \emph{arXiv preprint arXiv:2209.07989}, 2022{\natexlab{b}}.

\bibitem[Carion et~al.(2020)Carion, Massa, Synnaeve, Usunier, Kirillov, and Zagoruyko]{carion2020end}
Nicolas Carion, Francisco Massa, Gabriel Synnaeve, Nicolas Usunier, Alexander Kirillov, and Sergey Zagoruyko.
\newblock End-to-end object detection with transformers.
\newblock In \emph{Computer Vision--ECCV 2020: 16th European Conference, Glasgow, UK, August 23--28, 2020, Proceedings, Part I 16}, pp.\  213--229. Springer, 2020.

\bibitem[Chen et~al.(2022)Chen, Sima, Li, Zheng, Xu, Geng, Li, He, Shi, Qiao, and Yan]{chen2022persformer}
Li~Chen, Chonghao Sima, Yang Li, Zehan Zheng, Jiajie Xu, Xiangwei Geng, Hongyang Li, Conghui He, Jianping Shi, Yu~Qiao, and Junchi Yan.
\newblock Persformer: 3d lane detection via perspective transformer and the openlane benchmark.
\newblock In \emph{European Conference on Computer Vision (ECCV)}, 2022.

\bibitem[Chen et~al.(2023)Chen, Zhang, Wang, Wang, and Zhao]{chen2023futr3d}
Xuanyao Chen, Tianyuan Zhang, Yue Wang, Yilun Wang, and Hang Zhao.
\newblock Futr3d: A unified sensor fusion framework for 3d detection.
\newblock In \emph{Proceedings of the IEEE/CVF Conference on Computer Vision and Pattern Recognition}, pp.\  172--181, 2023.

\bibitem[Cheng et~al.(2022)Cheng, Wang, Chen, Zhang, Zhang, Huang, Zhang, and Liu]{cheng2022sparse}
Tianheng Cheng, Xinggang Wang, Shaoyu Chen, Wenqiang Zhang, Qian Zhang, Chang Huang, Zhaoxiang Zhang, and Wenyu Liu.
\newblock Sparse instance activation for real-time instance segmentation.
\newblock In \emph{Proceedings of the IEEE/CVF Conference on Computer Vision and Pattern Recognition}, pp.\  4433--4442, 2022.

\bibitem[Efrat et~al.(2020)Efrat, Bluvstein, Oron, Levi, Garnett, and Shlomo]{efrat20203d}
Netalee Efrat, Max Bluvstein, Shaul Oron, Dan Levi, Noa Garnett, and Bat~El Shlomo.
\newblock 3d-lanenet+: Anchor free lane detection using a semi-local representation.
\newblock \emph{arXiv preprint arXiv:2011.01535}, 2020.

\bibitem[Feng et~al.(2022)Feng, Guo, Tan, Xu, Wang, and Ma]{feng2022rethinking}
Zhengyang Feng, Shaohua Guo, Xin Tan, Ke~Xu, Min Wang, and Lizhuang Ma.
\newblock Rethinking efficient lane detection via curve modeling.
\newblock In \emph{Proceedings of the IEEE/CVF Conference on Computer Vision and Pattern Recognition}, pp.\  17062--17070, 2022.

\bibitem[Garnett et~al.(2019)Garnett, Cohen, Pe'er, Lahav, and Levi]{garnett20193d}
Noa Garnett, Rafi Cohen, Tomer Pe'er, Roee Lahav, and Dan Levi.
\newblock 3d-lanenet: end-to-end 3d multiple lane detection.
\newblock In \emph{Proceedings of the IEEE/CVF International Conference on Computer Vision}, pp.\  2921--2930, 2019.

\bibitem[Guo et~al.(2020)Guo, Chen, Zhao, Zhang, Miao, Wang, and Choe]{guo2020gen}
Yuliang Guo, Guang Chen, Peitao Zhao, Weide Zhang, Jinghao Miao, Jingao Wang, and Tae~Eun Choe.
\newblock Gen-lanenet: A generalized and scalable approach for 3d lane detection.
\newblock In \emph{European Conference on Computer Vision}, pp.\  666--681. Springer, 2020.

\bibitem[He et~al.(2016)He, Zhang, Ren, and Sun]{he2016deep}
Kaiming He, Xiangyu Zhang, Shaoqing Ren, and Jian Sun.
\newblock Deep residual learning for image recognition.
\newblock In \emph{Proceedings of the IEEE conference on computer vision and pattern recognition}, pp.\  770--778, 2016.

\bibitem[Hou et~al.(2019)Hou, Ma, Liu, and Loy]{hou2019learning}
Yuenan Hou, Zheng Ma, Chunxiao Liu, and Chen~Change Loy.
\newblock Learning lightweight lane detection cnns by self attention distillation.
\newblock In \emph{Proceedings of the IEEE/CVF international conference on computer vision}, pp.\  1013--1021, 2019.

\bibitem[Huang et~al.(2023)Huang, Shen, Huang, Ding, Dai, Han, Wang, and Liu]{huang2023anchor3dlane}
Shaofei Huang, Zhenwei Shen, Zehao Huang, Zi-han Ding, Jiao Dai, Jizhong Han, Naiyan Wang, and Si~Liu.
\newblock Anchor3dlane: Learning to regress 3d anchors for monocular 3d lane detection.
\newblock In \emph{Proceedings of the IEEE/CVF Conference on Computer Vision and Pattern Recognition}, pp.\  17451--17460, 2023.

\bibitem[Jang et~al.(2016)Jang, Gu, and Poole]{jang2016categorical}
Eric Jang, Shixiang Gu, and Ben Poole.
\newblock Categorical reparameterization with gumbel-softmax.
\newblock \emph{arXiv preprint arXiv:1611.01144}, 2016.

\bibitem[Jin et~al.(2022)Jin, Park, Jeong, Kwon, and Kim]{jin2022eigenlanes}
Dongkwon Jin, Wonhui Park, Seong-Gyun Jeong, Heeyeon Kwon, and Chang-Su Kim.
\newblock Eigenlanes: Data-driven lane descriptors for structurally diverse lanes.
\newblock In \emph{Proceedings of the IEEE/CVF Conference on Computer Vision and Pattern Recognition}, pp.\  17163--17171, 2022.

\bibitem[Kingma \& Ba(2014)Kingma and Ba]{kingma2014adam}
Diederik~P Kingma and Jimmy Ba.
\newblock Adam: A method for stochastic optimization.
\newblock \emph{arXiv preprint arXiv:1412.6980}, 2014.

\bibitem[Ko et~al.(2021)Ko, Lee, Azam, Munir, Jeon, and Pedrycz]{ko2021key}
Yeongmin Ko, Younkwan Lee, Shoaib Azam, Farzeen Munir, Moongu Jeon, and Witold Pedrycz.
\newblock Key points estimation and point instance segmentation approach for lane detection.
\newblock \emph{IEEE Transactions on Intelligent Transportation Systems}, 23\penalty0 (7):\penalty0 8949--8958, 2021.

\bibitem[Lang et~al.(2019)Lang, Vora, Caesar, Zhou, Yang, and Beijbom]{lang2019pointpillars}
Alex~H Lang, Sourabh Vora, Holger Caesar, Lubing Zhou, Jiong Yang, and Oscar Beijbom.
\newblock Pointpillars: Fast encoders for object detection from point clouds.
\newblock In \emph{Proceedings of the IEEE/CVF conference on computer vision and pattern recognition}, pp.\  12697--12705, 2019.

\bibitem[Lee et~al.(2017)Lee, Kim, Shin~Yoon, Shin, Bailo, Kim, Lee, Seok~Hong, Han, and So~Kweon]{lee2017vpgnet}
Seokju Lee, Junsik Kim, Jae Shin~Yoon, Seunghak Shin, Oleksandr Bailo, Namil Kim, Tae-Hee Lee, Hyun Seok~Hong, Seung-Hoon Han, and In~So~Kweon.
\newblock Vpgnet: Vanishing point guided network for lane and road marking detection and recognition.
\newblock In \emph{Proceedings of the IEEE international conference on computer vision}, pp.\  1947--1955, 2017.

\bibitem[Li et~al.(2022{\natexlab{a}})Li, Shi, Wang, and Cheng]{li2022reconstruct}
Chenguang Li, Jia Shi, Ya~Wang, and Guangliang Cheng.
\newblock Reconstruct from top view: A 3d lane detection approach based on geometry structure prior.
\newblock In \emph{Proceedings of the IEEE/CVF Conference on Computer Vision and Pattern Recognition}, pp.\  4370--4379, 2022{\natexlab{a}}.

\bibitem[Li et~al.(2019)Li, Li, Hu, and Yang]{li2019line}
Xiang Li, Jun Li, Xiaolin Hu, and Jian Yang.
\newblock Line-cnn: End-to-end traffic line detection with line proposal unit.
\newblock \emph{IEEE Transactions on Intelligent Transportation Systems}, 21\penalty0 (1):\penalty0 248--258, 2019.

\bibitem[Li et~al.(2022{\natexlab{b}})Li, Yu, Meng, Caine, Ngiam, Peng, Shen, Lu, Zhou, Le, et~al.]{li2022deepfusion}
Yingwei Li, Adams~Wei Yu, Tianjian Meng, Ben Caine, Jiquan Ngiam, Daiyi Peng, Junyang Shen, Yifeng Lu, Denny Zhou, Quoc~V Le, et~al.
\newblock Deepfusion: Lidar-camera deep fusion for multi-modal 3d object detection.
\newblock In \emph{Proceedings of the IEEE/CVF Conference on Computer Vision and Pattern Recognition}, pp.\  17182--17191, 2022{\natexlab{b}}.

\bibitem[Li et~al.(2022{\natexlab{c}})Li, Ge, Yu, Yang, Wang, Shi, Sun, and Li]{li2022bevdepth}
Yinhao Li, Zheng Ge, Guanyi Yu, Jinrong Yang, Zengran Wang, Yukang Shi, Jianjian Sun, and Zeming Li.
\newblock Bevdepth: Acquisition of reliable depth for multi-view 3d object detection.
\newblock \emph{arXiv preprint arXiv:2206.10092}, 2022{\natexlab{c}}.

\bibitem[Li et~al.(2022{\natexlab{d}})Li, Wang, Li, Xie, Sima, Lu, Qiao, and Dai]{li2022bevformer}
Zhiqi Li, Wenhai Wang, Hongyang Li, Enze Xie, Chonghao Sima, Tong Lu, Yu~Qiao, and Jifeng Dai.
\newblock Bevformer: Learning bird’s-eye-view representation from multi-camera images via spatiotemporal transformers.
\newblock In \emph{Computer Vision--ECCV 2022: 17th European Conference, Tel Aviv, Israel, October 23--27, 2022, Proceedings, Part IX}, pp.\  1--18. Springer, 2022{\natexlab{d}}.

\bibitem[Liang et~al.(2023)Liang, Zhou, Liu, and Wang]{liang2023clustseg}
James Liang, Tianfei Zhou, Dongfang Liu, and Wenguan Wang.
\newblock Clustseg: Clustering for universal segmentation.
\newblock \emph{arXiv preprint arXiv:2305.02187}, 2023.

\bibitem[Liang et~al.(2019)Liang, Yang, Chen, Hu, and Urtasun]{liang2019multi}
Ming Liang, Bin Yang, Yun Chen, Rui Hu, and Raquel Urtasun.
\newblock Multi-task multi-sensor fusion for 3d object detection.
\newblock In \emph{Proceedings of the IEEE/CVF Conference on Computer Vision and Pattern Recognition}, pp.\  7345--7353, 2019.

\bibitem[Liang et~al.(2022)Liang, Xie, Yu, Xia, Lin, Wang, Tang, Wang, and Tang]{liang2022bevfusion}
Tingting Liang, Hongwei Xie, Kaicheng Yu, Zhongyu Xia, Zhiwei Lin, Yongtao Wang, Tao Tang, Bing Wang, and Zhi Tang.
\newblock Bevfusion: A simple and robust lidar-camera fusion framework.
\newblock \emph{Advances in Neural Information Processing Systems}, 35:\penalty0 10421--10434, 2022.

\bibitem[Liao et~al.(2022)Liao, Chen, Wang, Cheng, Zhang, Liu, and Huang]{liao2022maptr}
Bencheng Liao, Shaoyu Chen, Xinggang Wang, Tianheng Cheng, Qian Zhang, Wenyu Liu, and Chang Huang.
\newblock Maptr: Structured modeling and learning for online vectorized hd map construction.
\newblock \emph{arXiv preprint arXiv:2208.14437}, 2022.

\bibitem[Lin et~al.(2017)Lin, Goyal, Girshick, He, and Doll{\'a}r]{lin2017focal}
Tsung-Yi Lin, Priya Goyal, Ross Girshick, Kaiming He, and Piotr Doll{\'a}r.
\newblock Focal loss for dense object detection.
\newblock In \emph{Proceedings of the IEEE international conference on computer vision}, pp.\  2980--2988, 2017.

\bibitem[Liu et~al.(2023{\natexlab{a}})Liu, Chen, Tan, Liu, Wang, Su, Li, Yao, Han, Ding, et~al.]{liu2023group}
Huan Liu, Qiang Chen, Zichang Tan, Jiang-Jiang Liu, Jian Wang, Xiangbo Su, Xiaolong Li, Kun Yao, Junyu Han, Errui Ding, et~al.
\newblock Group pose: A simple baseline for end-to-end multi-person pose estimation.
\newblock \emph{arXiv preprint arXiv:2308.07313}, 2023{\natexlab{a}}.

\bibitem[Liu et~al.(2018)Liu, Lehman, Molino, Petroski~Such, Frank, Sergeev, and Yosinski]{liu2018intriguing}
Rosanne Liu, Joel Lehman, Piero Molino, Felipe Petroski~Such, Eric Frank, Alex Sergeev, and Jason Yosinski.
\newblock An intriguing failing of convolutional neural networks and the coordconv solution.
\newblock \emph{Advances in neural information processing systems}, 31, 2018.

\bibitem[Liu et~al.(2021)Liu, Yuan, Liu, and Xiong]{liu2021end}
Ruijin Liu, Zejian Yuan, Tie Liu, and Zhiliang Xiong.
\newblock End-to-end lane shape prediction with transformers.
\newblock In \emph{Proceedings of the IEEE/CVF winter conference on applications of computer vision}, pp.\  3694--3702, 2021.

\bibitem[Liu et~al.(2022)Liu, Chen, Liu, Xiong, and Yuan]{liu2022learning}
Ruijin Liu, Dapeng Chen, Tie Liu, Zhiliang Xiong, and Zejian Yuan.
\newblock Learning to predict 3d lane shape and camera pose from a single image via geometry constraints.
\newblock In \emph{Proceedings of the AAAI Conference on Artificial Intelligence}, volume~36, pp.\  1765--1772, 2022.

\bibitem[Liu et~al.(2023{\natexlab{b}})Liu, Tang, Amini, Yang, Mao, Rus, and Han]{liu2023bevfusion}
Zhijian Liu, Haotian Tang, Alexander Amini, Xinyu Yang, Huizi Mao, Daniela~L Rus, and Song Han.
\newblock Bevfusion: Multi-task multi-sensor fusion with unified bird's-eye view representation.
\newblock In \emph{2023 IEEE International Conference on Robotics and Automation (ICRA)}, pp.\  2774--2781. IEEE, 2023{\natexlab{b}}.

\bibitem[Loshchilov \& Hutter(2016)Loshchilov and Hutter]{loshchilov2016sgdr}
Ilya Loshchilov and Frank Hutter.
\newblock Sgdr: Stochastic gradient descent with warm restarts.
\newblock \emph{arXiv preprint arXiv:1608.03983}, 2016.

\bibitem[Luo et~al.(2022{\natexlab{a}})Luo, Yan, Zheng, Zheng, Mei, Kun, Cui, and Li]{luo2022m}
Yueru Luo, Xu~Yan, Chaoda Zheng, Chao Zheng, Shuqi Mei, Tang Kun, Shuguang Cui, and Zhen Li.
\newblock M\^{} 2-3dlanenet: Multi-modal 3d lane detection.
\newblock \emph{arXiv preprint arXiv:2209.05996}, 2022{\natexlab{a}}.

\bibitem[Luo et~al.(2023)Luo, Zheng, Yan, Kun, Zheng, Cui, and Li]{luo2023latr}
Yueru Luo, Chaoda Zheng, Xu~Yan, Tang Kun, Chao Zheng, Shuguang Cui, and Zhen Li.
\newblock Latr: 3d lane detection from monocular images with transformer, 2023.

\bibitem[Luo et~al.(2022{\natexlab{b}})Luo, Zhou, Zhang, and Lu]{luo2022detr4d}
Zhipeng Luo, Changqing Zhou, Gongjie Zhang, and Shijian Lu.
\newblock Detr4d: Direct multi-view 3d object detection with sparse attention.
\newblock \emph{arXiv preprint arXiv:2212.07849}, 2022{\natexlab{b}}.

\bibitem[Luo et~al.(2024)Luo, Zhou, Pan, Zhang, Liu, Luo, Zhao, Liu, and Lu]{luo2024exploring}
Zhipeng Luo, Changqing Zhou, Liang Pan, Gongjie Zhang, Tianrui Liu, Yueru Luo, Haiyu Zhao, Ziwei Liu, and Shijian Lu.
\newblock Exploring point-bev fusion for 3d point cloud object tracking with transformer.
\newblock \emph{IEEE transactions on pattern analysis and machine intelligence}, 2024.

\bibitem[Ma et~al.(2022)Ma, Wang, Bai, Yang, Hou, Wang, Qiao, Yang, Manocha, and Zhu]{ma2022vision}
Yuexin Ma, Tai Wang, Xuyang Bai, Huitong Yang, Yuenan Hou, Yaming Wang, Yu~Qiao, Ruigang Yang, Dinesh Manocha, and Xinge Zhu.
\newblock Vision-centric bev perception: A survey.
\newblock \emph{arXiv preprint arXiv:2208.02797}, 2022.

\bibitem[Mallot et~al.(1991)Mallot, B{\"u}lthoff, Little, and Bohrer]{mallot1991inverse}
Hanspeter~A Mallot, Heinrich~H B{\"u}lthoff, JJ~Little, and Stefan Bohrer.
\newblock Inverse perspective mapping simplifies optical flow computation and obstacle detection.
\newblock \emph{Biological cybernetics}, 64\penalty0 (3):\penalty0 177--185, 1991.

\bibitem[Nedevschi et~al.(2004)Nedevschi, Schmidt, Graf, Danescu, Frentiu, Marita, Oniga, and Pocol]{nedevschi20043d}
Sergiu Nedevschi, Rolf Schmidt, Thorsten Graf, Radu Danescu, Dan Frentiu, Tiberiu Marita, Florin Oniga, and Ciprian Pocol.
\newblock 3d lane detection system based on stereovision.
\newblock In \emph{Proceedings. The 7th International IEEE Conference on Intelligent Transportation Systems (IEEE Cat. No. 04TH8749)}, pp.\  161--166. IEEE, 2004.

\bibitem[Neven et~al.(2018)Neven, De~Brabandere, Georgoulis, Proesmans, and Van~Gool]{neven2018towards}
Davy Neven, Bert De~Brabandere, Stamatios Georgoulis, Marc Proesmans, and Luc Van~Gool.
\newblock Towards end-to-end lane detection: an instance segmentation approach.
\newblock In \emph{2018 IEEE intelligent vehicles symposium (IV)}, pp.\  286--291. IEEE, 2018.

\bibitem[Oord et~al.(2018)Oord, Li, and Vinyals]{oord2018representation}
Aaron van~den Oord, Yazhe Li, and Oriol Vinyals.
\newblock Representation learning with contrastive predictive coding.
\newblock \emph{arXiv preprint arXiv:1807.03748}, 2018.

\bibitem[Pan et~al.(2017)Pan, Shi, Luo, Wang, and Tang]{pan2017spatial}
X~Pan, J~Shi, P~Luo, X~Wang, and X~Tang.
\newblock Spatial as deep: spatial cnn for traffic scene understanding. 2017.
\newblock \emph{Pan X Shi J Luo P Spatial As Deep: Spatial CNN for Traffic Scene Understanding}, 10, 2017.

\bibitem[Philion \& Fidler(2020)Philion and Fidler]{philion2020lift}
Jonah Philion and Sanja Fidler.
\newblock Lift, splat, shoot: Encoding images from arbitrary camera rigs by implicitly unprojecting to 3d.
\newblock In \emph{Computer Vision--ECCV 2020: 16th European Conference, Glasgow, UK, August 23--28, 2020, Proceedings, Part XIV 16}, pp.\  194--210. Springer, 2020.

\bibitem[Qu et~al.(2021)Qu, Jin, Zhou, Yang, and Zhang]{qu2021focus}
Zhan Qu, Huan Jin, Yang Zhou, Zhen Yang, and Wei Zhang.
\newblock Focus on local: Detecting lane marker from bottom up via key point.
\newblock In \emph{Proceedings of the IEEE/CVF Conference on Computer Vision and Pattern Recognition}, pp.\  14122--14130, 2021.

\bibitem[Radford et~al.(2021)Radford, Kim, Hallacy, Ramesh, Goh, Agarwal, Sastry, Askell, Mishkin, Clark, Krueger, and Sutskever]{radford2021learning}
Alec Radford, Jong~Wook Kim, Chris Hallacy, Aditya Ramesh, Gabriel Goh, Sandhini Agarwal, Girish Sastry, Amanda Askell, Pamela Mishkin, Jack Clark, Gretchen Krueger, and Ilya Sutskever.
\newblock Learning transferable visual models from natural language supervision, 2021.

\bibitem[Ren et~al.(2015)Ren, He, Girshick, and Sun]{ren2015faster}
Shaoqing Ren, Kaiming He, Ross Girshick, and Jian Sun.
\newblock Faster r-cnn: Towards real-time object detection with region proposal networks.
\newblock \emph{Advances in neural information processing systems}, 28, 2015.

\bibitem[Shi et~al.(2022)Shi, Li, and Ma]{shi2022pillarnet}
Guangsheng Shi, Ruifeng Li, and Chao Ma.
\newblock Pillarnet: Real-time and high-performance pillar-based 3d object detection.
\newblock In \emph{European Conference on Computer Vision}, pp.\  35--52. Springer, 2022.

\bibitem[Sindagi et~al.(2019)Sindagi, Zhou, and Tuzel]{sindagi2019mvx}
Vishwanath~A Sindagi, Yin Zhou, and Oncel Tuzel.
\newblock Mvx-net: Multimodal voxelnet for 3d object detection.
\newblock In \emph{2019 International Conference on Robotics and Automation (ICRA)}, pp.\  7276--7282. IEEE, 2019.

\bibitem[Sun et~al.(2020)Sun, Kretzschmar, Dotiwalla, Chouard, Patnaik, Tsui, Guo, Zhou, Chai, Caine, et~al.]{sun2020scalability}
Pei Sun, Henrik Kretzschmar, Xerxes Dotiwalla, Aurelien Chouard, Vijaysai Patnaik, Paul Tsui, James Guo, Yin Zhou, Yuning Chai, Benjamin Caine, et~al.
\newblock Scalability in perception for autonomous driving: Waymo open dataset.
\newblock In \emph{Proceedings of the IEEE/CVF conference on computer vision and pattern recognition}, pp.\  2446--2454, 2020.

\bibitem[Tabelini et~al.(2021{\natexlab{a}})Tabelini, Berriel, Paixao, Badue, De~Souza, and Oliveira-Santos]{tabelini2021keep}
Lucas Tabelini, Rodrigo Berriel, Thiago~M Paixao, Claudine Badue, Alberto~F De~Souza, and Thiago Oliveira-Santos.
\newblock Keep your eyes on the lane: Real-time attention-guided lane detection.
\newblock In \emph{Proceedings of the IEEE/CVF conference on computer vision and pattern recognition}, pp.\  294--302, 2021{\natexlab{a}}.

\bibitem[Tabelini et~al.(2021{\natexlab{b}})Tabelini, Berriel, Paixao, Badue, De~Souza, and Oliveira-Santos]{tabelini2021polylanenet}
Lucas Tabelini, Rodrigo Berriel, Thiago~M Paixao, Claudine Badue, Alberto~F De~Souza, and Thiago Oliveira-Santos.
\newblock Polylanenet: Lane estimation via deep polynomial regression.
\newblock In \emph{2020 25th International Conference on Pattern Recognition (ICPR)}, pp.\  6150--6156. IEEE, 2021{\natexlab{b}}.

\bibitem[Van~Gansbeke et~al.(2019)Van~Gansbeke, De~Brabandere, Neven, Proesmans, and Van~Gool]{van2019end}
Wouter Van~Gansbeke, Bert De~Brabandere, Davy Neven, Marc Proesmans, and Luc Van~Gool.
\newblock End-to-end lane detection through differentiable least-squares fitting.
\newblock In \emph{Proceedings of the IEEE/CVF International Conference on Computer Vision Workshops}, pp.\  0--0, 2019.

\bibitem[Wang et~al.(2021)Wang, Ma, Zhu, and Yang]{wang2021pointaugmenting}
Chunwei Wang, Chao Ma, Ming Zhu, and Xiaokang Yang.
\newblock Pointaugmenting: Cross-modal augmentation for 3d object detection.
\newblock In \emph{Proceedings of the IEEE/CVF Conference on Computer Vision and Pattern Recognition}, pp.\  11794--11803, 2021.

\bibitem[Wang et~al.(2022)Wang, Ma, Huang, Hui, Wang, Qian, and Zhang]{wang2022keypoint}
Jinsheng Wang, Yinchao Ma, Shaofei Huang, Tianrui Hui, Fei Wang, Chen Qian, and Tianzhu Zhang.
\newblock A keypoint-based global association network for lane detection.
\newblock In \emph{Proceedings of the IEEE/CVF Conference on Computer Vision and Pattern Recognition}, pp.\  1392--1401, 2022.

\bibitem[Wang et~al.(2023)Wang, Qin, Li, Li, Cao, and Xu]{wang2023bev}
Ruihao Wang, Jian Qin, Kaiying Li, Yaochen Li, Dong Cao, and Jintao Xu.
\newblock Bev-lanedet: An efficient 3d lane detection based on virtual camera via key-points.
\newblock In \emph{Proceedings of the IEEE/CVF Conference on Computer Vision and Pattern Recognition}, pp.\  1002--1011, 2023.

\bibitem[Wang et~al.(2018)Wang, Ren, and Qiu]{wang2018lanenet}
Ze~Wang, Weiqiang Ren, and Qiang Qiu.
\newblock Lanenet: Real-time lane detection networks for autonomous driving.
\newblock \emph{arXiv preprint arXiv:1807.01726}, 2018.

\bibitem[Wu et~al.(2018)Wu, Xiong, Yu, and Lin]{wu2018unsupervised}
Zhirong Wu, Yuanjun Xiong, Stella~X Yu, and Dahua Lin.
\newblock Unsupervised feature learning via non-parametric instance discrimination.
\newblock In \emph{Proceedings of the IEEE conference on computer vision and pattern recognition}, pp.\  3733--3742, 2018.

\bibitem[Xu et~al.(2020)Xu, Wang, Cai, Zhang, Liang, and Li]{xu2020curvelane}
Hang Xu, Shaoju Wang, Xinyue Cai, Wei Zhang, Xiaodan Liang, and Zhenguo Li.
\newblock Curvelane-nas: Unifying lane-sensitive architecture search and adaptive point blending.
\newblock In \emph{Computer Vision--ECCV 2020: 16th European Conference, Glasgow, UK, August 23--28, 2020, Proceedings, Part XV 16}, pp.\  689--704. Springer, 2020.

\bibitem[Xu et~al.(2022)Xu, Cai, Zhao, Zhang, Xu, Fu, and Xue]{xu2022rclane}
Shenghua Xu, Xinyue Cai, Bin Zhao, Li~Zhang, Hang Xu, Yanwei Fu, and Xiangyang Xue.
\newblock Rclane: Relay chain prediction for lane detection.
\newblock In \emph{European Conference on Computer Vision}, pp.\  461--477. Springer, 2022.

\bibitem[Yan et~al.(2022)Yan, Nie, Cai, Han, Xu, Yang, Ye, Fu, Mi, and Zhang]{yan2022once}
Fan Yan, Ming Nie, Xinyue Cai, Jianhua Han, Hang Xu, Zhen Yang, Chaoqiang Ye, Yanwei Fu, Michael~Bi Mi, and Li~Zhang.
\newblock Once-3dlanes: Building monocular 3d lane detection.
\newblock In \emph{Proceedings of the IEEE/CVF Conference on Computer Vision and Pattern Recognition}, pp.\  17143--17152, 2022.

\bibitem[Yang et~al.(2022)Yang, Chen, Miao, Li, Zhu, and Zhang]{yang2022deepinteraction}
Zeyu Yang, Jiaqi Chen, Zhenwei Miao, Wei Li, Xiatian Zhu, and Li~Zhang.
\newblock Deepinteraction: 3d object detection via modality interaction.
\newblock \emph{Advances in Neural Information Processing Systems}, 35:\penalty0 1992--2005, 2022.

\bibitem[Yao et~al.(2023)Yao, Yu, Wu, and Jia]{yao2023sparse}
Chengtang Yao, Lidong Yu, Yuwei Wu, and Yunde Jia.
\newblock Sparse point guided 3d lane detection.
\newblock In \emph{Proceedings of the IEEE/CVF International Conference on Computer Vision}, pp.\  8363--8372, 2023.

\bibitem[Yin et~al.(2020)Yin, Cheng, Wu, Song, Yu, and Niu]{yin2020fusionlane}
Ruochen Yin, Yong Cheng, Huapeng Wu, Yuntao Song, Biao Yu, and Runxin Niu.
\newblock Fusionlane: Multi-sensor fusion for lane marking semantic segmentation using deep neural networks.
\newblock \emph{IEEE Transactions on Intelligent Transportation Systems}, 23\penalty0 (2):\penalty0 1543--1553, 2020.

\bibitem[Yin et~al.(2021{\natexlab{a}})Yin, Zhou, and Krahenbuhl]{yin2021center}
Tianwei Yin, Xingyi Zhou, and Philipp Krahenbuhl.
\newblock Center-based 3d object detection and tracking.
\newblock In \emph{Proceedings of the IEEE/CVF conference on computer vision and pattern recognition}, pp.\  11784--11793, 2021{\natexlab{a}}.

\bibitem[Yin et~al.(2021{\natexlab{b}})Yin, Zhou, and Kr{\"a}henb{\"u}hl]{yin2021multimodal}
Tianwei Yin, Xingyi Zhou, and Philipp Kr{\"a}henb{\"u}hl.
\newblock Multimodal virtual point 3d detection.
\newblock \emph{Advances in Neural Information Processing Systems}, 34:\penalty0 16494--16507, 2021{\natexlab{b}}.

\bibitem[Yoo et~al.(2020)Yoo, Kim, Kim, and Choi]{yoo20203d}
Jin~Hyeok Yoo, Yecheol Kim, Jisong Kim, and Jun~Won Choi.
\newblock 3d-cvf: Generating joint camera and lidar features using cross-view spatial feature fusion for 3d object detection.
\newblock In \emph{Computer Vision--ECCV 2020: 16th European Conference, Glasgow, UK, August 23--28, 2020, Proceedings, Part XXVII 16}, pp.\  720--736. Springer, 2020.

\bibitem[Yu et~al.(2023)Yu, Tao, Xie, Lin, Liang, Wang, Chen, Hao, Wang, and Liang]{yu2023benchmarking}
Kaicheng Yu, Tang Tao, Hongwei Xie, Zhiwei Lin, Tingting Liang, Bing Wang, Peng Chen, Dayang Hao, Yongtao Wang, and Xiaodan Liang.
\newblock Benchmarking the robustness of lidar-camera fusion for 3d object detection.
\newblock In \emph{Proceedings of the IEEE/CVF Conference on Computer Vision and Pattern Recognition}, pp.\  3187--3197, 2023.

\bibitem[Yu et~al.(2022)Yu, Wang, Qiao, Collins, Zhu, Adam, Yuille, and Chen]{yu2022k}
Qihang Yu, Huiyu Wang, Siyuan Qiao, Maxwell Collins, Yukun Zhu, Hartwig Adam, Alan Yuille, and Liang-Chieh Chen.
\newblock k-means mask transformer.
\newblock In \emph{European Conference on Computer Vision}, pp.\  288--307. Springer, 2022.

\bibitem[Zhang et~al.(2021{\natexlab{a}})Zhang, Cai, Yan, Feng, et~al.]{zhang2021direct}
Jianfeng Zhang, Yujun Cai, Shuicheng Yan, Jiashi Feng, et~al.
\newblock Direct multi-view multi-person 3d pose estimation.
\newblock \emph{Advances in Neural Information Processing Systems}, 34:\penalty0 13153--13164, 2021{\natexlab{a}}.

\bibitem[Zhang et~al.(2021{\natexlab{b}})Zhang, Li, Gao, Jin, and Li]{zhang2021channel}
Xinyu Zhang, Zhiwei Li, Xin Gao, Dafeng Jin, and Jun Li.
\newblock Channel attention in lidar-camera fusion for lane line segmentation.
\newblock \emph{Pattern Recognition}, 118:\penalty0 108020, 2021{\natexlab{b}}.

\bibitem[Zheng et~al.(2021)Zheng, Fang, Zhang, Tang, Yang, Liu, and Cai]{zheng2021resa}
Tu~Zheng, Hao Fang, Yi~Zhang, Wenjian Tang, Zheng Yang, Haifeng Liu, and Deng Cai.
\newblock Resa: Recurrent feature-shift aggregator for lane detection.
\newblock In \emph{Proceedings of the AAAI Conference on Artificial Intelligence}, volume~35, pp.\  3547--3554, 2021.

\bibitem[Zhou \& Tuzel(2018)Zhou and Tuzel]{zhou2018voxelnet}
Yin Zhou and Oncel Tuzel.
\newblock Voxelnet: End-to-end learning for point cloud based 3d object detection.
\newblock In \emph{Proceedings of the IEEE conference on computer vision and pattern recognition}, pp.\  4490--4499, 2018.

\bibitem[Zhu et~al.(2020)Zhu, Su, Lu, Li, Wang, and Dai]{zhu2020deformable}
Xizhou Zhu, Weijie Su, Lewei Lu, Bin Li, Xiaogang Wang, and Jifeng Dai.
\newblock Deformable detr: Deformable transformers for end-to-end object detection.
\newblock \emph{arXiv preprint arXiv:2010.04159}, 2020.

\end{thebibliography}
\bibliographystyle{iclr2024_conference}

\clearpage

\appendix
\section{Appendix}

\setcounter{table}{0}
\setcounter{figure}{0}

\begin{wraptable}{rhp}{0.5\columnwidth}
\vspace{-8mm}
\caption{\textbf{Model complexity.} FPS is evaluated on a single V100 GPU.}
\label{tab:fps} 
\begin{center}
\vspace{2mm}
{
\scalebox{.8}{
\begin{tabular}{c|c|r|c}
\toprule[1.2pt]
 \textbf{Model} & \textbf{Backbone} & \textbf{FPS} & \textbf{F1}\\ 
        \hline
        \addlinespace
        \multicolumn{1}{l|}{PersFormer} & Efficient-B7 & 11.67 & 36.5 \\
        \multicolumn{1}{l|}{PersFormer} & Res50 & 9.48 & 43.2 \\
        \multicolumn{1}{l|}{M$^2$-3DLaneNet} & Efficient-B7 & 6.48 & 48.2 \\
        \multicolumn{1}{l|}{Anchor3DLane} & Efficient-B3 & 3.07 & 34.9 \\
        \multicolumn{1}{l|}{Anchor3DLane} & Res18 & 3.45 & 32.8 \\
        \multicolumn{1}{l|}{LATR} & Res50 & 13.34 & 54.0 \\
        \cline{1-4}
        \addlinespace
        \multicolumn{1}{l|}{\modelname-Tiny} & Res18, PillarNet18 & 13.49 & 60.9 \\
        \multicolumn{1}{l|}{\modelname-Base} & Res34, PillarNet34 & 8.82 & 63.5 \\
        \multicolumn{1}{l|}{\modelname-Large} & Res50, PillarNet34 & 6.18 & 65.2 \\
\bottomrule[1.2pt]
\end{tabular}}
}
\end{center}
\hspace{0.2\textwidth}
\vspace{-4mm}
\end{wraptable}

\subsection{Model Complexity}
\vspace{-2mm}
As stated in our main paper, DV-3DLane achieves SoTA performance, and its lite version also surpasses all previous methods in terms of F1 score and localization errors, while achieving an impressive FPS of 13.49.
In this section, we study the model complexity, as shown in~\Cref{tab:fps}. Our base model achieves a competitive FPS of 8.82 while maintaining a strong F1 score of 63.5. 
Notably, our tiny version excels with an FPS of 13.49, along with a notable F1 score of 60.9.

\subsection{Scenario studies}
\begin{figure*}[pth!]
\centering
   \includegraphics[width=0.98\linewidth]{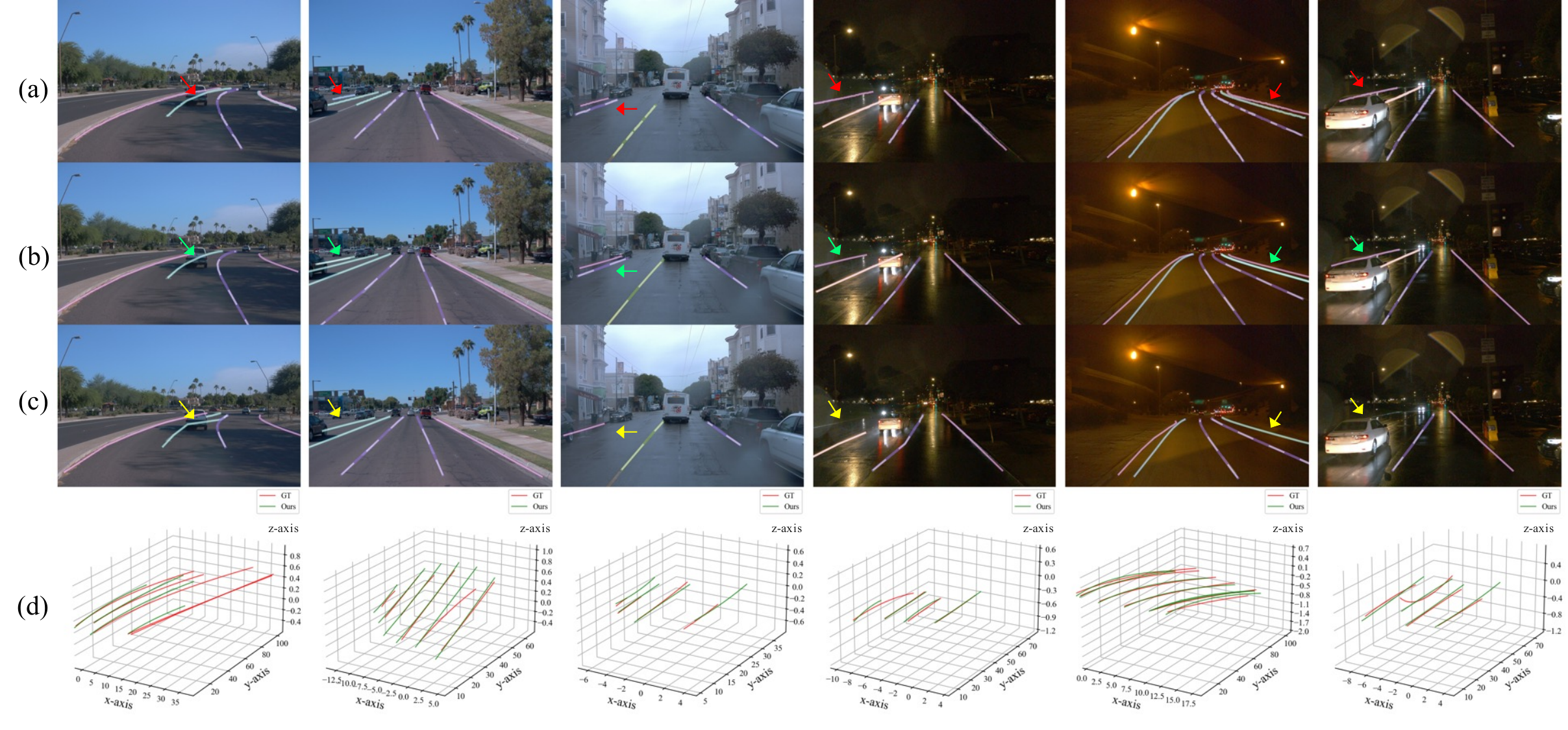}
   \vspace{-4mm}
\caption{\textbf{More Results.} Rows (a), (b), (c) show projections of 3D lanes from the \textcolor{gt}{ground truth} (GT), \textcolor{Green}{\modelname}, and \textcolor{prev_sota}{LATR}~\cite{luo2023latr}, with differences highlighted by colored arrows. Row (d) compares GT (\textcolor{gt}{red}) and our prediction (\textcolor{Green}{green}) in 3D. Best viewed in color and zoom in for details.}
\label{fig:supp_qua_results}
\end{figure*}

\vspace{-2mm}
Additionally, we comprehensively evaluated DV-3DLane across \textit{diverse scenarios} within OpenLane. As depicted in~\Cref{tab:main_result_scene}, our method consistently outperforms all previous approaches across all six challenging scenarios by a large margin. Visualizations are provided in~\Cref{fig:supp_qua_results}. 
Overall, these results reveal the effectiveness of our design.

\vspace{-4mm}
\begin{table*}[h!]
\small
    \centering
    \setlength{\extrarowheight}{2.2pt}
    \caption{Comparison with other 3D lane detection methods on the OpenLane validation dataset. $\dagger$ denotes that the results are obtained using their provided models.}
    \label{tab:main_result_scene}
    \vspace{2mm}
    \resizebox{\textwidth}{!}{
    \begin{tabular}{c|c|c|c|c|c|c|c|c|c|c}
    \Xhline{1.8pt}
     \multirow{2}{*}{\textbf{\textit{Dist.}}} 
     & \multirow{2}{*}{\textbf{Methods}} 
     & \multirow{2}{*}{\textbf{Backbone}} 
     & \multirow{2}{*}{\textbf{Modality}} 
     & \multirow{2}{*}{\textbf{All}} 
     & \textbf{Up \&}  & \multirow{2}{*}{\textbf{Curve}} 
     & \textbf{Extreme} & \multirow{2}{*}{\textbf{Night}} 
     & \multirow{2}{*}{\textbf{Intersection}} 
     & \textbf{Merge} \\
      & & & & &  \textbf{Down} & & \textbf{Weather} & & & \textbf{\& Split} \\
    \Xhline{0.5pt}
    \addlinespace[1.5pt]
    \multirow{12}{*}{{\rotatebox{90}{\small{\textbf{1.5 m}}}}} 
    & \multicolumn{1}{l|}{3DLaneNet~\cite{garnett20193d}} & VGG-16 & C & 44.1 & 40.8 & 46.5 & 47.5 & 41.5 & 32.1 & 41.7 \\
    & \multicolumn{1}{l|}{GenLaneNet~\cite{guo2020gen}} & ERFNet & C & 32.3 & 25.4 & 33.5 & 28.1 & 18.7 & 21.4 & 31.0 \\
    & \multicolumn{1}{l|}{PersFormer~\cite{chen2022persformer}} & EffNet-B7 & C & 50.5 & 42.4 & 55.6 & 48.6 & 46.6 & 40.0 & 50.7 \\
    & \multicolumn{1}{l|}{Anchor3DLane~\cite{huang2023anchor3dlane}$^\dagger$} & EffNet-B3 & C & 52.8 & 48.5 & 50.7 & 56.9 & 43.6 & 48.5 & 50.7 \\
    & \multicolumn{1}{l|}{M$^2$-3DLaneNet~\cite{luo2022m}} & EffNet-B7 & C+L & 55.5 & 53.4 & 60.7 & 56.2 & 51.6 & 43.8 & 51.4 \\
    & \multicolumn{1}{l|}{PersFormer~\cite{chen2022persformer}} & ResNet-50 & C & 52.7 & 46.4 & 57.9 & 52.9 & 47.2 & 41.6 & 51.4 \\
    & \multicolumn{1}{l|}{LATR~\cite{luo2023latr}} & ResNet-50 & C & \underline{61.9} & \underline{55.2} & \underline{68.2} & \underline{57.1} & \underline{55.4} & \underline{52.3} & \underline{61.5} \\
    & \multicolumn{1}{l|}{Anchor3DLane~\cite{huang2023anchor3dlane}$^\dagger$} & ResNet-18 & C & 50.7 & 45.3 & 53.7 & 48.5 & 51.6 & 45.3 & 48.5 \\
    & \multicolumn{1}{l|}{\cellcolor{mygray}{\modelname-Tiny}} & \cellcolor{mygray}{ResNet-18} & \cellcolor{mygray}{C+L} & \cellcolor{mygray}{63.4} & \cellcolor{mygray}{59.9} & \cellcolor{mygray}{69.8} & \cellcolor{mygray}{62.2} & \cellcolor{mygray}{58.8} & \cellcolor{mygray}{53.5} & \cellcolor{mygray}{60.6} \\
    & \multicolumn{1}{l|}{\cellcolor{mygray}{\modelname-Base}} & \cellcolor{mygray}{ResNet-34} & \cellcolor{mygray}{C+L} & \cellcolor{mygray}{65.4} & \cellcolor{mygray}{60.9} & \cellcolor{mygray}{\textbf{72.1}} & \cellcolor{mygray}{64.5} & \cellcolor{mygray}{61.3} & \cellcolor{mygray}{55.5} & \cellcolor{mygray}{61.6} \\
    &  \multicolumn{1}{l|}{\cellcolor{mygray}{\modelname-Large}} & \cellcolor{mygray}{ResNet-50} & \cellcolor{mygray}{C+L} & \cellcolor{mygray}{\textbf{66.8}} & \cellcolor{mygray}{\textbf{61.1}} & \cellcolor{mygray}{71.5} & \cellcolor{mygray}{\textbf{64.9}} & \cellcolor{mygray}{\textbf{63.2}} & \cellcolor{mygray}{\textbf{58.6}} & \cellcolor{mygray}{\textbf{62.8}} \\
    \cline{2-11}
    \addlinespace[1.2pt]
    & \emph{Improvement} & - & -
    & \textcolor{impblue}{\textit{$\uparrow$4.9}}
    & \textcolor{impblue}{\textit{$\uparrow$5.9}}
    & \textcolor{impblue}{\textit{$\uparrow$3.9}}
    & \textcolor{impblue}{\textit{$\uparrow$7.8}}
    & \textcolor{impblue}{\textit{$\uparrow$7.8}}
    & \textcolor{impblue}{\textit{$\uparrow$6.3}}
    & \textcolor{impblue}{\textit{$\uparrow$1.3}}
    \\
    \Xhline{0.6pt}
    \addlinespace[1.5pt]
    \multirow{10}{*}{{\rotatebox{90}{\small{\textbf{0.5 m}}}}}
    & \multicolumn{1}{l|}{PersFormer~\cite{chen2022persformer}} & EffNet-B7 & C & 36.5 & 26.8 & 36.9 & 33.9 & 34.0 & 28.5 & 37.4 \\
    & \multicolumn{1}{l|}{Anchor3DLane~\cite{huang2023anchor3dlane}$^\dagger$} & EffNet-B3 & C & 34.9 & 28.3 & 31.8 & 30.7 & 32.2 & 29.9 & 33.9 \\
    &  \multicolumn{1}{l|}{M$^2$-3DLaneNet~\cite{luo2022m}} & EffNet-B7 & C+L & 48.2 & 40.7 & 48.2 & \underline{49.8} & \underline{46.2} & 38.7 & 44.2 \\
    &\multicolumn{1}{l|}{PersFormer~\cite{chen2022persformer}} & ResNet-50 & C & 43.2 & 36.3 & 42.4 & 45.4 & 39.3 & 32.9 & 41.7 \\
    & \multicolumn{1}{l|}{LATR~\cite{luo2023latr}} & ResNet-50 & C & \underline{54.0} & \underline{44.9} & \underline{56.2} & 47.6 & \underline{46.2} & \underline{45.5} & \underline{55.6} \\
    & \multicolumn{1}{l|}{Anchor3DLane~\cite{huang2023anchor3dlane}$^\dagger$} & ResNet-18 & C & 32.8 & 26.5 & 27.6 & 31.2 & 30.0 & 28.1 & 31.7 \\
    & \multicolumn{1}{l|}{\cellcolor{mygray}{\modelname-Tiny}} & \cellcolor{mygray}{ResNet-18} & \cellcolor{mygray}{C+L} & \cellcolor{mygray}{60.9} & \cellcolor{mygray}{56.9} & \cellcolor{mygray}{65.9} & \cellcolor{mygray}{60.0} & \cellcolor{mygray}{56.8} & \cellcolor{mygray}{50.7} & \cellcolor{mygray}{57.6} \\
    &  \multicolumn{1}{l|}{\cellcolor{mygray}{\modelname-Base}} & \cellcolor{mygray}{ResNet-34} & \cellcolor{mygray}{C+L} & \cellcolor{mygray}{63.5} & \cellcolor{mygray}{58.6} & \cellcolor{mygray}{\textbf{69.3}} & \cellcolor{mygray}{62.4} & \cellcolor{mygray}{59.9} & \cellcolor{mygray}{53.9} & \cellcolor{mygray}{59.3} \\
    &  \multicolumn{1}{l|}{\cellcolor{mygray}{\modelname-Large}} & \cellcolor{mygray}{ResNet-50} & \cellcolor{mygray}{C+L} & \cellcolor{mygray}{\textbf{65.2}} & \cellcolor{mygray}{\textbf{59.1}} & \cellcolor{mygray}{69.2} & \cellcolor{mygray}{\textbf{63.0}} & \cellcolor{mygray}{\textbf{62.0}} & \cellcolor{mygray}{\textbf{56.9}} & \cellcolor{mygray}{\textbf{60.5}} \\
    \cline{2-11}
    \addlinespace[1.2pt]
    & \emph{Improvement} & - & -
    & \textcolor{impblue}{\textit{$\uparrow$11.2}}
    & \textcolor{impblue}{\textit{$\uparrow$14.2}}
    & \textcolor{impblue}{\textit{$\uparrow$13.1}}
    & \textcolor{impblue}{\textit{$\uparrow$13.2}}
    & \textcolor{impblue}{\textit{$\uparrow$15.8}}
    & \textcolor{impblue}{\textit{$\uparrow$11.4}}
    & \textcolor{impblue}{\textit{$\uparrow$4.9}}
    \\
    \Xhline{1.2pt}
    \end{tabular}
    }
\end{table*}

\subsection{Robustness}
To investigate the robustness of our model amid calibration noise, given that perfect calibration is not always viable in real-world settings, we conduct experiments incorporating diverse levels of calibration noise to understand the model's performance under noisy conditions.

\paragraph{Noise settings:}
Following the methodology of ~\cite{yu2023benchmarking}, we introduce two noise settings: \textbf{Noise (N)} and \textbf{Stronger Noise (SN)}.
In `Noise (N),' we introduce random rotations within [1$^\circ$, 5$^{\circ}$] and translations within [0.5cm, 1.0cm] to the calibration. For `Stronger Noise (SN),' these parameters are doubled to reflect stronger calibration disturbances.

\paragraph{Results without training noise:}
We first test our model, which has \textit{not} been trained with additional noise under these noisy conditions. 
The results, presented in the first row of~\Cref{tab:noise}, show a notable performance decline as the noise intensity increases.
Specifically, under a probability setting of 0.7, the performance deteriorates from 63.5 to 32.4/31.4 in the 'Noise'/'Stronger Noise' settings.

\paragraph{Enhancing robustness via training:}
To enhance robustness, we incorporate calibration noise during the training phase. This strategy substantially mitigates the performance degradation caused by noisy calibration, as shown in the second and third rows of~\Cref{tab:noise}. 

\paragraph{Comparative analysis:}
Compared to the baseline (first row), we can observe that training with calibration noise significantly strengthens the robustness of our model. It effectively maintains comparable results under noisy calibration conditions.
Additionally, the model trained with 'Stronger Noise' exhibits greater robustness compared to the one trained with less intense noise, underscoring the benefits of this training strategy.

\begin{table*}[ph!] \centering
    \newcommand{\Frst}[1]{\textcolor{red}{\textbf{#1}}}
    \newcommand{\Scnd}[1]{\textcolor{blue}{\textbf{#1}}}
    \caption{Impact of noise on calibration parameters. We set two noise levels in the experiments, ``\textbf{Noise (N)}" and ``\textbf{Stronger Noise (SN)}." In the \textbf{Train} column, ``-" denotes \textit{no} noise is added during the training phase. ``Prob" denotes the probability of adding the corresponding noise into the training/eval phases. Each result group consists of \texttt{F1-score / Accuracy.}}
    \label{tab:noise}
    \vspace{2mm}
    \resizebox{\textwidth}{!}{
    \small
    \begin{tabular}{cc||c|ccc|ccc}
    \toprule
     \multicolumn{2}{c||}{\textbf{Train}} & \multicolumn{7}{c}{\textbf{Eval}} \\
     \multicolumn{2}{c||}{\multirow{2}{*}{@noise (\textbf{N/SN})}} &  \multirow{1}{*||}{} & \multicolumn{3}{c|}{\textbf{+ Noise (N)}} & \multicolumn{3}{c}{\textbf{+ Stronger Noise (SN)}}  \\
     & & Prob=0.0 & Prob=0.3 & Prob=0.5 & Prob=0.7 & Prob=0.3 & {Prob=0.5} &  Prob=0.7\\
    \midrule
     - &
      Prob=0.0 & 63.5 / 92.4 & 52.2 / 89.9 & 40.9 / 85.6  & 32.4 / 82.5 &  52.0 / 89.0 & 40.3 / 83.4 & 31.4 / 79.2 \\
    \textbf{N} & Prob=0.3 & 63.0 / 93.1 & 62.5 / 92.9 & 62.0 / 92.9 & 61.5 / 92.9  & 62.2 / 92.9
                  & 61.5 / 92.9 & 60.8 / 92.7 \\
    \textbf{SN} &
     Prob=0.3   &  63.4 / 92.5  & 62.8 / 92.4 & 62.3 / 92.3 & 61.8 / 92.2 & 62.7 / 92.4 & 62.1 / 92.2 & 61.7 / 92.2 \\
    \bottomrule
\end{tabular}
}
\end{table*}

\subsection{Effect of Dual-view}
Apart from studying the impact of multiple modalities, we conducted experiments on the OpenLane-1K dataset to analyze the effect of the dual views, providing a comprehensive understanding of our approach. 
As shown in~\Cref{tab:views}, we conducted two sets of experiments: \textbf{1)} Using single modality and single view. \textbf{2)} Using single modality but dual views.

In the first set of experiments, rows \#1 and \#2 present the performance using individual modalities.

In the second set of experiments: 
\begin{itemize}
    \item  For the image branch experiment, we adopt a strategy similar to BEVFormer~\cite{li2022bevformer}, utilizing deformable attention to transform image features into BEV features. Then, we apply our dual-view decoder upon this, and the outcomes are illustrated in row \#3 of~\Cref{tab:views}.
    \item For the LiDAR branch experiment, we project LiDAR point cloud features onto the 2D image plane to generate perspective-view features. The results of this approach are presented in row \#4 of~\Cref{tab:views}.
\end{itemize}

The results in~\Cref{tab:views} underscore that the dual-view representation significantly enhances the performance of baseline models in single-modal scenarios (comparing \#1 with \#3 and \#2 with \#4). This improvement confirms the effectiveness of our dual-view approach in learning 3D lane detection. Most notably, the combination of image and LiDAR modalities, coupled with our dual-view representation, achieves the best results, as shown in row \#5. This synergy of modalities underlines the superiority of our proposed method.

\begin{table}[h!]\centering
    \footnotesize
    \renewcommand{\arraystretch}{1.2}
    \caption{Comparison of single and dual-view approaches on OpenLane-1K dataset with 0.5m setting.} 
    \vspace{2mm}
    \label{tab:views}
\begin{tabular}{c| c | c | c | c | c | c | c } 
\toprule[1.2pt]
 \multirow{2}{*}{\# Line} & \multirow{2}{*}{Inputs}  & \multirow{2}{*}{View} & \multirow{2}{*}{Backbone} & \multirow{2}{*}{F1} & \multirow{2}{*}{Acc.}  & {X error (m)} & {Z error (m)} \\
 & & & & & & \multicolumn{1}{c|}{\makebox[0pt][c]{\hspace{-1.7em}\textit{near}\hspace{1em}}$\vert$\makebox[0pt][c]{\hspace{3.2em}\textit{far}\hspace{0.5em}}} & \multicolumn{1}{c}{\makebox[0pt][c]{\hspace{-1.7em}\textit{near}\hspace{1em}}$\vert$\makebox[0pt][c]{\hspace{3.2em}\textit{far}\hspace{0.5em}}}\\
 \Xhline{0.5pt}
 \addlinespace[1.2pt]
 \#1 & Image & PV & Res34 & 52.9 & 90.3 & 0.173 $\vert$ 0.212 & 0.069 $\vert$ 0.098 \\
 \#2 & LiDAR & BEV & PillarNet34 & 54.1 & 84.4 & 0.282 $\vert$ 0.191 & 0.096 $\vert$ 0.123 \\
 \#3 & Image & Dual Views & Res34  & 54.3 & 91.5 & 0.165 $\vert$ 0.200 & 0.067 $\vert$ 0.094 \\
 \#4 & LiDAR & Dual Views & PillarNet34 & 55.3 & 87.9 & 0.156 $\vert$ 0.143 & 0.031 $\vert$ 0.050 \\
 \#5 & DV-3DLane & Dual Views & Res34+PillarNet34 & 63.5 & 92.4 & 0.090 $\vert$ 0.102 & 0.031 $\vert$ 0.053 \\
\bottomrule[1.2pt]
\end{tabular}
\end{table}

\subsection{Effect of Bidirectional Feature Fusion.}

To validate the effectiveness of this strategy, we compare the performance of our method with the other three fusion design choices, as shown in~\Cref{tab:early_fusion}, where ``Cam" means only image features in \uv branch, and ``LiDAR" denotes only point features in \bev branch. ``L$\rightarrow$C" denotes the LiDAR to camera fusion for \uv branch, and conversely, ``C$\rightarrow$L" denotes the camera to LiDAR fusion for \bev branch. 
It shows that the absence of fusion leads to the poorest performance (\#1).
Further, employing one-way fusion, either from camera to LiDAR (\#2) or LiDAR to camera (\#3), results in 1.2\% and 1.1\% improvements, respectively \wrt non-fusion (\#1).
Remarkably, our bidirectional fusion (\#4) yields the highest performance, a 2.8\% gain in F1.
This improvement highlights the efficacy of our strategy in effectively leveraging multi-modal features in both \uv and \bev spaces.

\begin{table}[h]
\begin{center}
    \renewcommand{\arraystretch}{1.2}
    \caption{Effect of bidirectional feature fusion.
    }
    \label{tab:early_fusion}
    \vspace{2mm}
    {
    \begin{tabular}{c|c|c|c|c}
    \toprule[1.2pt]
    \multirow{2}{*}{\# Line} & \multirow{2}{*}{Methods}  & \multirow{2}{*}{F1} & {X error (m)} & {Z error (m)} \\
     & & & \multicolumn{1}{c|}{\makebox[0pt][c]{\hspace{-1.7em}\textit{near}\hspace{1em}}$\vert$\makebox[0pt][c]{\hspace{3.2em}\textit{far}\hspace{0.5em}}} & \multicolumn{1}{c}{\makebox[0pt][c]{\hspace{-1.7em}\textit{near}\hspace{1em}}$\vert$\makebox[0pt][c]{\hspace{3.2em}\textit{far}\hspace{0.5em}}}\\
    \hline
    \addlinespace[2pt]
    \#1 & \multicolumn{1}{l|}{Cam~~ \& LiDAR} & 67.9 & 0.133 $\vert$ 0.157 & 0.060 $\vert$ 0.083 \\
    \#2 & \multicolumn{1}{l|}{Cam~~ \& C$\rightarrow$L} & 69.1 & 0.135 $\vert$ 0.151 & 0.060 $\vert$ 0.081 \\
    \#3 & \multicolumn{1}{l|}{L$\rightarrow$C \& LiDAR} & 69.0 & 0.130 $\vert$ 0.156 & 0.059 $\vert$ 0.078 \\
    \#4 & \multicolumn{1}{l|}{L$\rightarrow$C \& C$\rightarrow$L} & 70.7 & 0.123 $\vert$ 0.146 & 0.058 $\vert$ 0.078 \\
    \toprule[1.2pt]
    \end {tabular}
    }
\end{center}
\end{table}

\subsection{Image Branch on Apollo}
Table.~\ref{tab:apollo} illustrates the results of our image branch on the Apollo dataset~\cite{guo2020gen}, compared with existing methods. 

\begin{table*}[t!]
\vspace{-12cm}
\begin{center}
\small
\caption{\textbf{Results on Apollo 3D Synthetic dataset.} ``Image-Branch" denotes the image branch of our DV-3DLane.
}
\label{tab:apollo}
\resizebox{1.0\linewidth}{!}{
  \footnotesize
  \begin{tabular}{c|clcllll}
  \toprule
  \multicolumn{1}{c}{} &\multicolumn{1}{c}{} & \multicolumn{1}{c}{} & \multicolumn{1}{c}{} & \multicolumn{2}{c}{{X error (m)} $\downarrow$} & \multicolumn{2}{c}{{Z error (m)} $\downarrow$} \\
  \cmidrule{5-6} \cmidrule{7-8} 
  \multicolumn{1}{c}{\multirow{-2}{*}{{Scene}}} &
  \multicolumn{1}{c}{\multirow{-2}{*}{{Methods}}} & \multicolumn{1}{l}{\multirow{-2}{*}{{F1} $\uparrow$}} & \multicolumn{1}{l}{\multirow{-2}{*}{{AP} $\uparrow$}} & \multicolumn{1}{l}{\textit{near}} & \multicolumn{1}{l}{\,\textit{far}} & \multicolumn{1}{l}{\textit{near}} & \multicolumn{1}{l}{\,\textit{far}} \\ \hline\hline
\addlinespace[1.5pt]
\multirow{10}{*}{Balanced Scene} & 
\multicolumn{1}{l}{3DLaneNet~\cite{garnett20193d}} & 86.4 & 89.3 \quad\quad\quad & 0.068 & 0.477 & 0.015 & {0.202} \\
& \multicolumn{1}{l}{Gen-LaneNet~\cite{guo2020gen}} & 88.1 & 90.1 \quad\quad\quad &  0.061 & 0.496 & 0.012 & 0.214 \\
& \multicolumn{1}{l}{CLGo~\cite{liu2022learning}} & 91.9 & 94.2 \quad\quad\quad &  0.061 & 0.361 & 0.029 & 0.250 \\
& \multicolumn{1}{l}{PersFormer~\cite{chen2022persformer}} & 92.9 & - \quad\quad\quad &  0.054 & 0.356 & 0.010 & 0.234 \\
& \multicolumn{1}{l}{GP~\cite{li2022reconstruct}} & 91.9 & 93.8 \quad\quad\quad & {0.049} & 0.387 & {0.008} & {0.213} \\
& \multicolumn{1}{l}{CurveFormer~\cite{bai2022curveformer}} & {95.8} & {97.3} \quad\quad\quad & 0.078 & {0.326} & 0.018 & 0.219 \\
& \multicolumn{1}{l}{Anchor3DLane~\cite{huang2023anchor3dlane}} & {95.6} & {97.2} \quad\quad\quad & 0.052 & {0.306} & 0.015 & 0.223 \\
& \multicolumn{1}{l}{LATR~\cite{luo2023latr}} & {96.8} & {97.9} \quad\quad\quad & 0.022 & {0.253} & 0.007 & 0.202 \\
& \multicolumn{1}{l}{Image-Branch (Ours)} & {96.4} & {97.6} \quad\quad\quad & 0.046 & {0.299} & 0.016 & 0.213 \\

\Xhline{0.6pt}
\addlinespace[1.5pt]
\multirow{10}{*}{Rare Subset} & 
\multicolumn{1}{l}{3DLaneNet~\cite{garnett20193d}} & 74.6 & 72.0 \quad\quad\quad & 0.166 & 0.855 & 0.039 & {0.521} \\
& \multicolumn{1}{l}{Gen-LaneNet~\cite{guo2020gen}} & 78.0 & 79.0 \quad\quad\quad &  0.139 & 0.903 & 0.030 & 0.539 \\
& \multicolumn{1}{l}{CLGo~\cite{liu2022learning}} & 86.1 & 88.3 \quad\quad\quad &  0.147 & {0.735} & 0.071 & 0.609 \\
& \multicolumn{1}{l}{PersFormer~\cite{chen2022persformer}} & 87.5 & - \quad\quad\quad & {0.107} & 0.782 & 0.024 & 0.602 \\
& \multicolumn{1}{l}{GP~\cite{li2022reconstruct}} & 83.7 & 85.2 \quad\quad\quad & 0.126 & 0.903 & {0.023} & 0.625 \\
& \multicolumn{1}{l}{CurveFormer~\cite{bai2022curveformer}} & {95.6} & {97.1} \quad\quad\quad & 0.182 & 0.737 & 0.039 & 0.561 \\
& \multicolumn{1}{l}{Anchor3DLane~\cite{huang2023anchor3dlane}} & {94.4} & {96.9} \quad\quad\quad & 0.094 & {0.693} & 0.027 & 0.579 \\
& \multicolumn{1}{l}{LATR~\cite{luo2023latr}} & {96.1} & {97.3} \quad\quad\quad & 0.050 & {0.600} & 0.015 & 0.532 \\
& \multicolumn{1}{l}{Image-Branch (Ours)} & {95.6} & {97.2} \quad\quad\quad & 0.071 & {0.664} & 0.025 & 0.568 \\
\Xhline{0.6pt}
\addlinespace[1.5pt]
\multirow{10}{*}{Visual Variations} 
& 
\multicolumn{1}{l}{3DLaneNet~\cite{garnett20193d}} & 74.9 & 72.5 \quad\quad\quad & 0.115 & 0.601 & 0.032 & {0.230} \\
& \multicolumn{1}{l}{Gen-LaneNet~\cite{guo2020gen}} & 85.3 & 87.2 \quad\quad\quad & 0.074 & 0.538 & 0.015 & 0.232 \\
& \multicolumn{1}{l}{CLGo~\cite{liu2022learning}} & 87.3 & 89.2 \quad\quad\quad & 0.084 & 0.464 & 0.045 & 0.312 \\
& \multicolumn{1}{l}{PersFormer~\cite{chen2022persformer}} & 89.6 \quad\quad\quad & - \quad\quad\quad& 0.074 & 0.430 & 0.015 & 0.266 \\
& \multicolumn{1}{l}{GP~\cite{li2022reconstruct}} & 89.9 & 92.1 \quad\quad\quad & {0.060} & 0.446 & {0.011 } & 0.235 \\
& \multicolumn{1}{l}{CurveFormer~\cite{bai2022curveformer}} & {90.8} & {93.0} \quad\quad\quad & 0.125 & {0.410} & 0.028 & 0.254 \\
& \multicolumn{1}{l}{Anchor3DLane~\cite{huang2023anchor3dlane}} & {91.4} & {93.6} \quad\quad\quad & 0.068 & {0.367} & 0.020 & 0.232 \\
& \multicolumn{1}{l}{LATR~\cite{luo2023latr}} & {95.1} & {96.6} \quad\quad\quad & 0.045 & {0.315} & 0.016 & 0.228 \\
& \multicolumn{1}{l}{Image-Branch (Ours)} & {91.3} & {93.4} \quad\quad\quad & 0.095 & {0.417} & 0.040 & 0.320 \\
\bottomrule  
\end{tabular}}

\end{center}
\vspace{-3mm}
\end{table*}

\end{document}